\documentclass[letterpaper,journal]{IEEEtran}
\usepackage{amsmath,amsfonts}
\usepackage{algorithm}
\usepackage{array}
\usepackage{subfigure}
\usepackage{textcomp}
\usepackage{stfloats}
\usepackage{url}
\usepackage{verbatim}
\usepackage{graphicx}
\usepackage{cite}
\usepackage{amssymb}
\usepackage{colortbl}
\usepackage{xcolor}
\usepackage{mathtools}
\usepackage{microtype}
\usepackage{float}
\usepackage{bm}
\usepackage{amsthm}
\usepackage{caption}
\usepackage{subcaption}
\usepackage{algpseudocode}
\usepackage{threeparttable}
\usepackage{pifont}
\usepackage{enumitem}
\usepackage{bbding}
\usepackage{fontawesome}
\usepackage{algorithmicx}
\usepackage{rotating}

\usepackage{booktabs}
\usepackage{multirow}
\usepackage{makecell}
\usepackage{siunitx}
\usepackage{adjustbox}
\usepackage{graphicx}

\usepackage{amsmath,amsfonts,bm}

\def\eqref#1{equation~\ref{#1}}

\def\1{\bm{1}}

\def\rvx{{\mathbf{x}}}

\DeclareMathAlphabet{\mathsfit}{\encodingdefault}{\sfdefault}{m}{sl}
\SetMathAlphabet{\mathsfit}{bold}{\encodingdefault}{\sfdefault}{bx}{n}

\newcommand{\Ls}{\mathcal{L}}

\newcommand{\softmax}{\mathrm{softmax}}

\DeclareMathOperator*{\argmax}{arg\,max}
\DeclareMathOperator*{\argmin}{arg\,min}

\newcommand{\rot}[1]{\rotatebox[origin=c]{60}{\textbf{#1}}}
\newcommand{\gain}[1]{\textcolor{gray}{#1}}

\definecolor{lightgray}{rgb}{0.9, 0.9, 0.9}
\definecolor{lightblue}{rgb}{0.8, 0.9, 1.0}
\definecolor{lightgreen}{rgb}{0.9, 1.0, 0.9}
\definecolor{lavender}{rgb}{0.9, 0.9, 0.98}
\definecolor{brightgreen}{RGB}{102, 255, 0}
\definecolor{tamegray}{gray}{0.94}
\definecolor{deepgreen}{RGB}{0,128,0}
\newcommand{\imp}[1]{\textcolor{deepgreen}{\scriptsize$^{\,+#1}$}}

\usepackage{xspace}
\makeatletter
\DeclareRobustCommand\onedot{\futurelet\@let@token\@onedot}
\def\@onedot{\ifx\@let@token.\else.\null\fi\xspace}

\makeatother

\begin{document}

\title{TAME: Test-Time Adversarial Prompt Tuning via Mixture-of-Experts for Vision-Language Models}

\author{%
Xin~Wang, Yixu~Wang, Jiaming~Zhang, Ruofan~Wang, Jiaqi~Yu, Kai~Chen, Jingjing~Chen,\\
Xingjun~Ma, and Yu-Gang~Jiang,~\IEEEmembership{Fellow,~IEEE}%
\thanks{This work has been submitted to the IEEE for possible publication. Copyright may be transferred without notice, after which this version may no longer be accessible.}%
}

\maketitle

\begin{abstract}
Large-scale pre-trained Vision-Language models (VLMs), such as CLIP, exhibit strong zero-shot generalization, yet remain highly vulnerable to imperceptible adversarial perturbations, raising serious safety concerns for open-world deployment. To enhance robustness without requiring downstream task-specific retraining, we propose \emph{TAME}, a novel test-time defense.
Building upon our prior Test-Time Adversarial Prompt Tuning (TAPT), TAME introduces an architectural reformulation by replacing TAPT's single adaptive prompt with an input-conditioned Mixture-of-Experts (MoE) framework, enabling more expressive and adaptive defense.
Specifically, TAME maintains a bank of learnable expert prompts and employs an input-dependent routing mechanism to aggregate a customized prompt mixture for each unlabeled test sample at inference time. This test-time defense mechanism is driven by three unsupervised objectives: (1) multi-view prediction entropy minimization, (2) layer-wise alignment of visual token statistics to precomputed clean and adversarial reference distributions, and (3) MoE regularization for balanced expert utilization and prompt diversity. We evaluated TAME on 11 benchmark datasets, including ImageNet and 10 additional zero-shot datasets. The results show that TAME improves the zero-shot adversarial robustness of the original CLIP by at least 49.1\% under AutoAttack while largely preserving generalization on clean samples. TAME also consistently outperforms existing adversarial prompt tuning methods across multiple prompt designs, yielding an average robustness gain of at least 30.2\%.
\end{abstract}


\section{Introduction}
\label{sec:intro}

Vision-Language Models (VLMs), pre-trained on web-scale image-text datasets, have become a foundation for zero-shot visual recognition and cross-modal understanding across a wide range of applications, including computer vision~\cite{radford2021learning,jia2021scaling,zhang2023multi}, medical image analysis~\cite{huang2023visual,wang2022medclip,min2022lossless,min2020web}, and embodied robotics~\cite{ahn2022can,shridhar2022cliport,khandelwal2022simple}. By leveraging large-scale cross-modal supervision, VLMs such as CLIP exhibit remarkable transferability to downstream tasks without task-specific tuning. However, despite their strong zero-shot generalization, VLMs remain highly vulnerable to adversarial examples, where imperceptible perturbations can disrupt image-text alignment and severely degrade performance~\cite{szegedy2013intriguing,madry2017towards,dong2018boosting,zhang2022towards,zhao2024evaluating,wang2026openrt,wang2025freezevla,ma2026safety,gao2025imperceptible,chen2025evolve}. Such vulnerability raises serious safety and reliability concerns for real-world deployment in open environments.

Existing defenses for adversarially robust VLMs are largely dominated by train-time defense paradigms, such as adversarial training~\cite{madry2017towards,zhang2019theoretically,gan2020large,wang2024revisiting} on predefined downstream tasks. Although effective under predefined task settings, these methods typically rely on task-specific training data and labels, which contradict the original zero-shot ability of pre-trained VLMs~\cite{su2018robustness, pedraza2021relationship}. In realistic deployment, however, the space of downstream tasks is effectively unbounded and often unknown in advance. Therefore, \emph{robustness evaluated only on predefined tasks is insufficient}; instead, the key challenge is to achieve zero-shot adversarial robustness, maintaining reliable performance under adversarial attacks on previously unseen tasks and datasets without downstream retraining.

\begin{figure}
    \centering
    \includegraphics[width=1\linewidth]{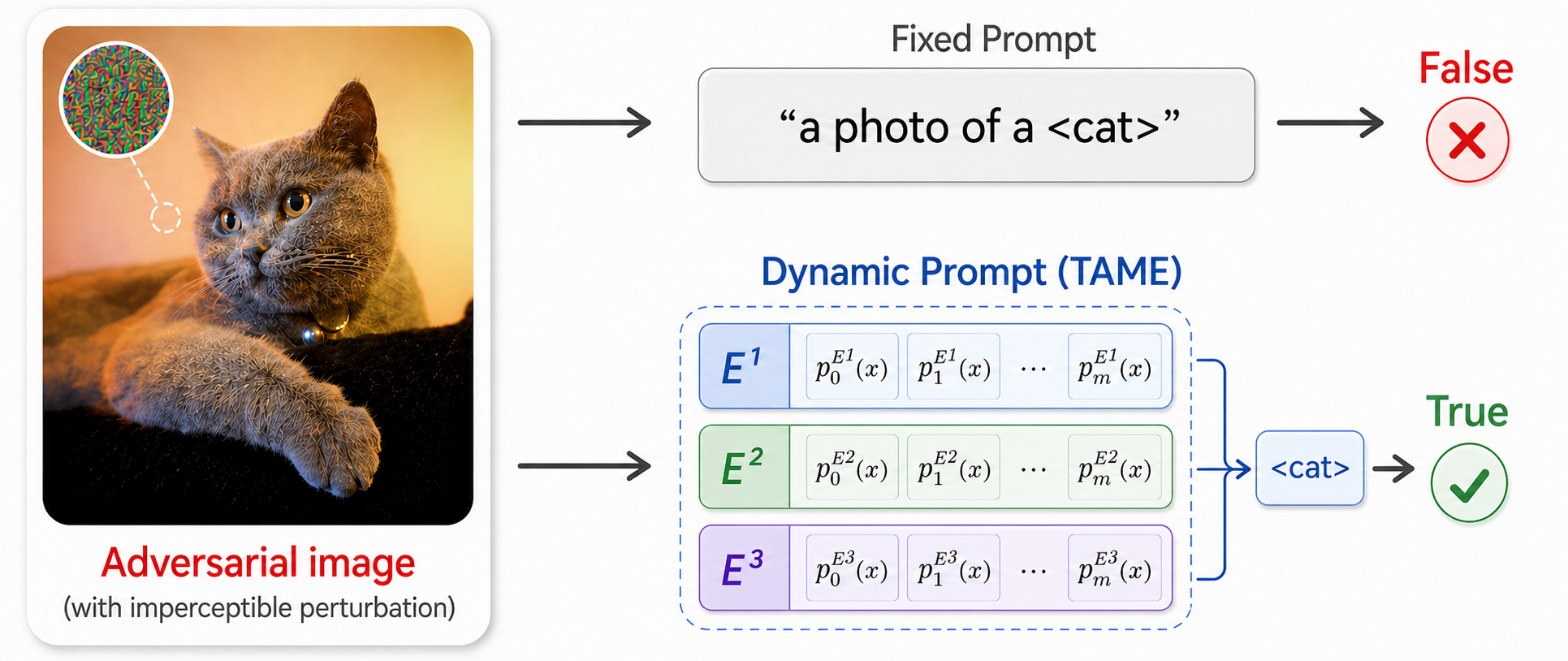}
    \caption{Inference with different prompts. \textbf{Top}: inference with hand-crafted prompts fails to recognize the class `cat'; \textbf{Bottom}: Inference with test-time adversarial mixture-of-experts optimized for each image produces accurate recognitions.}
    \label{fig1}
\end{figure}

To address these limitations, our preliminary work, Test-Time Adversarial Prompt Tuning (TAPT)~\cite{wang2025tapt}, introduced a test-time defense strategy that adapts prompts for each test sample while keeping the VLM backbone frozen. By dynamically realigning the adversarial image embedding with its corresponding text embedding during inference, TAPT improves zero-shot adversarial robustness without requiring task-specific downstream training data~\cite{zhang2023adversarial,li2024one,luo2024adversarial,zhou2024few}. 
However, TAPT remains fundamentally constrained by its single-prompt architecture.
(1) Limited expressive capacity. A single prompt imposes an inherent representation bottleneck and is often insufficient to compensate for the diverse and highly nonlinear distortions caused by different adversarial attacks.
(2) Limited adaptation space. TAPT follows a single adaptation pathway and therefore lacks a mechanism to selectively recruit complementary prompt behaviors according to the attack pattern, image content, and cross-modal discrepancy of each test sample.
(3) Limited architectural flexibility. TAPT formulates test-time defense as the optimization of one prompt instance, which inherently prevents the architecture from modeling multiple defensive behaviors in a structured manner. These limitations indicate a structural mismatch between single-prompt adaptation and test-time defense. Since adversarial shifts are input-dependent, TAPT's single prompt trajectory is insufficient to capture diverse defensive behaviors. This motivates test-time mixture adaptation, where multiple expert prompts are dynamically aggregated for each test sample to enable robust inference.

In this work, we propose \textbf{TAME}, a novel Test-Time Adversarial Mixture-of-Experts framework that reformulates TAPT's single-prompt adaptation as an input-conditioned Multimodal Mixture-of-Prompts (MMoP) formulation. Instead of optimizing one prompt per test sample, TAME maintains a bank of learnable expert prompts and employs a lightweight routing to aggregate a customized prompt mixture for each unlabeled input at inference time. This design turns the prompt from a single correction vector into a structured expert space, substantially increasing the expressive capacity of test-time adversarial prompt tuning, enabling sample-wise specialization, and supporting diverse prompt designs. As illustrated in Figure~\ref{fig1}, such adaptive expert composition yields more reliable recognition under adversarial perturbations than hand-crafted prompts. Concretely, TAME is optimized at test time with three unsupervised objectives: (1) \emph{multi-view prediction entropy minimization}, which encourages confident and consistent predictions across stochastic augmentations; (2) \emph{adversarial-clean alignment}, which aligns visual token statistics with precomputed adversarial and clean reference distributions; and (3) \emph{MoE regularization}, which promotes balanced expert utilization and prompt diversity. Notably, the entire adversarial adaptation procedure is performed at test time, requires no task-specific annotations, and keeps the VLM backbone frozen.

We conducted extensive evaluations on 11 benchmark datasets, including ImageNet and 10 additional zero-shot datasets, under both white-box and black-box adversarial attacks. The results demonstrate that TAME consistently outperforms our initial TAPT as well as other state-of-the-art baselines. Moreover, the consistent improvements achieved by MoE variants across three representative prompt designs (including Visual Only, V-L Joint, and V-L Independent) validate the generality and extensibility of the proposed TAME framework. Beyond empirical gains, our analysis reveals that TAME improves robust inference by dynamically composing complementary expert prompts, thereby expanding the prompt adaptation space and enabling per-sample correction of adversarial misalignment. These results confirm the effectiveness of TAME as a general test-time defense framework for robust inference, substantially improving zero-shot adversarial robustness while largely preserving clean generalization.

As an extension of our conference paper~\cite{wang2025tapt}, which introduced the concept of \emph{Test-Time Adversarial Prompt Tuning}, this work makes the following significant contributions:
\begin{itemize}
    \item We propose \emph{TAME}, a Test-Time Adversarial Mixture-of-Experts framework for robust zero-shot VLM inference. Building upon our original TAPT, TAME advances test-time defense from single-prompt adaptation to structured mixture adaptation, maintaining a bank of learnable expert prompts and dynamically routing expert prompts for each test sample to improve zero-shot adversarial robustness.

    \item We introduce a unified multimodal mixture-of-prompts formulation for Visual Only, V-L Joint, and V-L Independent designs. This defense mechanism is optimized via a mixture-aware unsupervised objective that integrates multi-view consistency, adversarial-clean alignment, and MoE regularization, requiring no task-specific annotations.

    \item We conduct extensive experiments on 11 datasets under white-box and black-box attacks, demonstrating that TAME consistently outperforms our TAPT as well as other state-of-the-art baselines. Specifically, TAME achieves zero-shot adversarial robustness of 54.1\% with ViT-B/16 and 49.1\% with ViT-B/32 under AutoAttack, respectively.
\end{itemize}

\section{Related Work}
\label{sec:related}

\subsection{Adversarial Attacks on Pre-trained VLMs}  
Adversarial attacks on pre-trained VLMs are broadly categorized as white-box or black-box attacks according to the threat model. In white-box attacks, the attacker has full access to the model parameters and can directly compute adversarial gradients~\cite{madry2017towards, zhang2022towards, zhou2023advclip, ma2024imbalanced}. Black-box attacks, on the other hand, restrict the attacker to querying model outputs~\cite{xie2019improving, zhao2024evaluating, lu2023set, he2023sa, wang2024transferable, yin2024vlattack, fang2024one, zhang2024universal}. Traditional single-modal attacks designed for vision models, such as PGD~\cite{madry2017towards}, DI~\cite{xie2019improving}, and AutoAttack~\cite{croce2020reliable}, can be used directly to attack the image encoders of pre-trained VLMs. Several recent multi-modal attacks have targeted pre-trained VLMs by simultaneously exploiting vulnerabilities in both their image and text encoders. For example, Co-Attack~\cite{zhang2022towards} pioneered white-box multi-modal attacks, perturbing both image and text modalities concurrently. SGA~\cite{lu2023set} extended Co-Attack to the black-box setting, improving the transferability of multi-modal adversarial examples. Building on this, SA-Attack~\cite{he2023sa} enhances cross-modal transferability by introducing data augmentations to both original and adversarial inputs. VLP-Attack~\cite{wang2023exploring} improves transferability by generating adversarial texts and images using contrastive loss. To overcome SGA's limitations, TMM~\cite{wang2024transferable} introduces modality-consistency and discrepancy features through attention-based and orthogonal-guided perturbations. VLATTACK~\cite{yin2024vlattack} further crafts adversarial examples by combining image and text perturbations at both single-modal and multi-modal levels.

\begin{figure*}[htbp]
    \centering
    \includegraphics[width=1\linewidth]{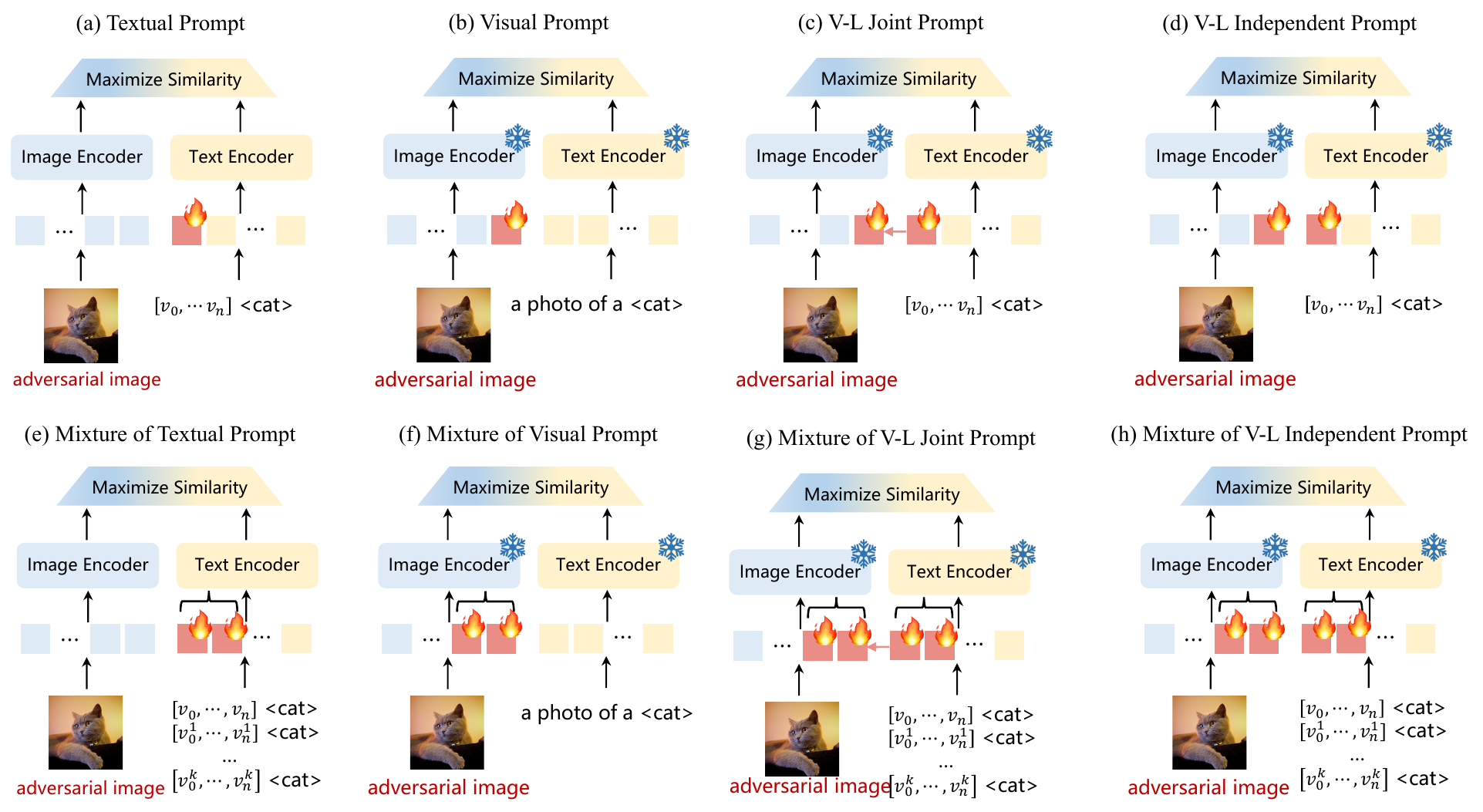}
    \caption{Illustration of adversarial prompt tuning designs. (a)--(d) show single-prompt designs, including Textual Prompts, Visual Prompts, V-L Joint Prompts, and V-L Independent Prompt; (e)--(h) show their corresponding MoE variants.}
    \label{fig2}
\end{figure*}

\subsection{Adversarial Defenses for Pre-trained VLMs} 
Adversarial training/tuning is a widely used defense strategy for pre-trained VLMs, with existing methods generally classified into adversarial contrastive tuning~\cite{wang2024revisiting, mao2023understanding, schlarmannrobust, wang2024pre, zhou2024revisiting, wang2024advqdet} and adversarial prompt tuning~\cite{zhang2023adversarial, li2024one, zhou2024few, hussein2024promptsmooth}. Adversarial contrastive tuning focuses on enhancing the adversarial robustness of the backbone model. For example, TeCoA~\cite{mao2023understanding} examines the impact of fine-tuning and visual prompt tuning on the zero-shot adversarial robustness of VLMs. FARE~\cite{schlarmannrobust} improves CLIP's image encoder through unsupervised adversarial fine-tuning, enhancing adversarial robustness in models like LLaVA and OpenFlamingo without retraining. PMG-AFT~\cite{wang2024pre} introduces an auxiliary branch to boost zero-shot adversarial robustness, while MMCoA~\cite{zhou2024revisiting} investigates VLM vulnerabilities to multimodal attacks. In contrast, AdvPT~\cite{zhang2023adversarial} and APT~\cite{li2024one} offer an efficient approach to bolster the adversarial robustness of VLMs by tuning only the textual prompts without altering the model parameters. FAP~\cite{zhou2024few} further refines APT by balancing cross-modal consistency between benign and adversarial inputs. MixPrompt~\cite{fan2024mixprompt} simultaneously enhances the generalizability and adversarial robustness of VLPs by employing conditional APT. APD~\cite{luo2024adversarial} improves CLIP’s robustness through online prompt distillation between teacher and student multi-modal prompts. While these methods are all training-time defense methods that need to pre-tune the prompt for a specific downstream task, in this work we propose a novel test-time defense method that optimizes the prompt on the fly during inference and is task-agnostic.

\subsection{Mixture-of-Experts} 
The Mixture of Experts (MoE) architecture, which trains multiple specialized subnetworks and uses a gating mechanism to select the most relevant expert for a given input, has been effectively adapted for prompt learning in VLMs. This Mixture-of-Prompts (MoP) approach addresses the limitation that a single prompt is often insufficient to handle diverse data distributions. Several methods have explored this paradigm.
PLOT~\cite{chen2022plot} learns a comprehensive set of prompts to capture varied category characteristics, utilizing optimal transport to align visual and textual features for improved few-shot recognition. Similarly, LASP~\cite{bulat2023lasp} employs multiple prompts but regularizes the tuning process using features from grouped manual templates. More dynamic approaches, such as MoCoOp~\cite{du2025mixture}, introduce a router network to select and combine the top-K most suitable soft prompts for each input image, enhancing adaptability to diverse styles or perturbations. Further research has reinforced the benefits of this strategy. MoP~\cite{wang2024one} demonstrates that optimizing a mixture of prompts leads to superior generalization across a range of tasks. Focusing on efficiency, SMoP~\cite{choi2023smop} introduces a sparse mixture-of-prompts framework. The strategy has also been tailored for security, with methods like AMPT~\cite{zhao2025enhancing} using a prompt mixture to specifically enhance robustness against adversarial attacks.

\subsection{Test-Time Adaptation}  
Test-time adaptation (TTA) methods enhance the generalization performance of pre-trained models by adapting to individual test samples or batches, addressing distribution shifts between training (source) and testing (target) data. 
Early TTA approaches~\cite{schneider2020improving, nado2020evaluating} focused on addressing domain shift by updating a model's batch normalization statistics to match those of the incoming test data. Building on this, TENT~\cite{wangtent} optimizes batch normalization layers by minimizing the entropy of predictions for each test batch, thereby increasing the model's confidence. MEMO~\cite{zhang2022memo} extends this concept by minimizing entropy over multiple augmented views of a single test input. To handle continuously evolving data streams, methods like CoTTA~\cite{wang2022continual} and EATA~\cite{niu2022efficient} were developed to enable adaptation to continuously test distributions. More recently, methods like TPT~\cite{shu2022test}, PromptAlign~\cite{abdul2024align}, and MTA~\cite{zanella2024test} have focused on prompt learning exclusively at test time to ensure consistent predictions across various augmented views of a test sample. Recent advances in test-time defense have yielded particularly impressive results in defending VLMs against adversarial attacks. Our previous work, TAPT~\cite{wang2025tapt}, introduced an efficient paradigm to adaptively find a robust prompt that aligns an adversarial image with pre-computed distributions. Concurrently, R-TPT~\cite{sheng2025r} enhances robustness using pointwise entropy minimization and a reliability-based ensembling strategy, while TTC~\cite{xing2025clip} presents a training-free defense that leverages CLIP's vision encoder to mitigate adversarial perturbations at inference.

However, most existing defenses rely on a single-prompt assumption, optimizing one prompt per input or task. Since adversarial perturbations are diverse and input-specific, a single prompt often lacks sufficient capacity to handle a wide range of attacks. This limitation motivates our TAME framework, which replaces the single-prompt bottleneck with  the mixture-of-prompts. At test time, TAME employs a lightweight gating network to synthesize an image-conditioned prompt from a small set of experts. This MoE design is broadly compatible with common prompt designs and yields consistent performance gains across different models and datasets.

\section{Preliminaries}
\subsection{Threat Model}
We consider a white-box threat model in which the adversary has full knowledge of the target model's architecture and parameters, enabling direct perturbation of test images via adversarial gradients prior to inference. The defender, the model owner, is permitted to employ any defense strategy to mitigate such adversarial attacks. In particular, we focus on safeguarding CLIP zero-shot inference, a setting in which the defender has access to neither task-specific training data nor annotations from the downstream application.

\subsection{Revisiting CLIP}
Let $\mathcal{I}$ denote the CLIP image encoder parameterized by $\theta_{\mathcal{I}}$, and let $\mathcal{T}$ denote the text encoder parameterized by $\theta_{\mathcal{T}}$. Consider a $K$-class classification task in which each image $\rvx$ is associated with a class label rendered in the template $\texttt{"a photo of a <class>"}$. Given a clean sample $\rvx \in [0, 1]^d$ and a target CLIP model $\{\mathcal{I},\mathcal{T}\}$, a white-box adversarial attack aims to craft an adversarial example $\rvx'$ that maximizes the model loss:
\begin{equation}
    \rvx' = \argmax_{\| \rvx' - \rvx \|_{\infty} \leq \epsilon} \Ls(\mathcal{I}(\rvx'), \mathcal{T}(y)),
\end{equation}
where $\Ls(\cdot)$ denotes the contrastive loss between the image and text embeddings, and $\epsilon$ specifies the $\ell_\infty$ perturbation budget.

\begin{algorithm}[t!]
    \caption{TAME}
    \label{algorithm}
    \begin{algorithmic}[1]
    \State {\bfseries Input:} input image $\rvx$, image encoder $\mathcal{I}$, text encoder $\mathcal{T}$, augmentation function $\mathcal{A}$, entropy threshold $\tau$, pre-computed $\mathcal{D}_{\text{public}}$ statistics $\{\mu_{\text{adv}}, \sigma^2_{\text{adv}}, \mu_{\text{clean}}, \sigma^2_{\text{clean}}\}$, MoE weights $\{\lambda_{\text{bal}}, \lambda_{\text{div}}\}$
    \State {\bfseries Output:} learnable prompts $\widetilde{\bm{P}}$

    \State \textbf{1. Initialize Mixture of Adversarial Prompts}:
    \State \quad $\widetilde{\bm{P}} \gets \text{APT}(\mathcal{D}_{\text{public}}, \epsilon; \mathcal{I}, \mathcal{T})$

    \State \textbf{2. Multi-View Entropy-Based Sample Selection:}
    \State \quad Select the top $\tau$ entropy from $\mathcal{A}(\rvx)$ to form $\mathcal{H}_\tau(\rvx)$

    \State \textbf{3. Compute Multi-View Entropy Loss:}
    \State \quad \scalebox{0.95}{$\Ls_{\text{entropy}} = -\sum\limits_{i=1}^{K} \Tilde{p}(y_i|\mathcal{H}_\tau(\rvx),\widetilde{\bm{P}}) \log \Tilde{p}(y_i|\mathcal{H}_\tau(\rvx),\widetilde{\bm{P}})$}

    \State \textbf{4. Compute Current Embedding Statistics:}
    \State \quad \textbf{for} each layer $l$ in $\mathcal{I}$ \textbf{do}
        \State \qquad $\mu_l = \text{Mean}(\mathcal{I}_l(\hat{\rvx}, \widetilde{\bm{P}}))$ for $\hat{\rvx}$ in $\mathcal{H}_\tau(\rvx)$
        \State \qquad $\sigma_l^2 = \text{Var}(\mathcal{I}_l(\hat{\rvx}, \widetilde{\bm{P}}))$ for $\hat{\rvx}$ in $\mathcal{H}_\tau(\rvx)$
    \State \quad \textbf{end for}

    \State \textbf{5. Adversarial-Clean Embedding Alignment:}
    \State \quad \scalebox{0.95}{$\Ls_{\text{adv}} = \dfrac{1}{L} \sum\limits_{l=1}^L \left( \left\| \mu_l - \mu_{\text{adv}, l} \right\|_1 + \left\| \sigma^2_l - \sigma^2_{\text{adv}, l} \right\|_1 \right)$}
    \State \quad \scalebox{0.95}{$\Ls_{\text{clean}} = \dfrac{1}{L} \sum\limits_{l=1}^L \left( \left\| \mu_l - \mu_{\text{clean}, l} \right\|_1 + \left\| \sigma^2_l - \sigma^2_{\text{clean}, l} \right\|_1 \right)$}

    \State \textbf{6. MoE Regularization:}
    \State \quad $\Ls_{\text{MoE}} = \sum\limits_{\ell\in\mathcal{L}_{\text{MoE}}}\!\big(\lambda_{\text{bal}}\Ls^{l}_{\text{bal}} + \lambda_{\text{div}}\Ls^{l}_{\text{div}}\big)$

    \State \textbf{7. Optimize Prompts:}
    \State \quad $\Ls_{\text{TAME}} = \Ls_{\text{entropy}} + \alpha \Ls_{\text{adv}} + (1-\alpha) \Ls_{\text{clean}} + \Ls_{\text{MoE}}$
    \State \quad Optimize $\widetilde{\bm{P}} \leftarrow$ Minimize $\Ls_{\text{TAME}}$
    
    \end{algorithmic}
\end{algorithm}

\begin{figure*}[htbp]
    \centering
    \includegraphics[width=0.8\linewidth]{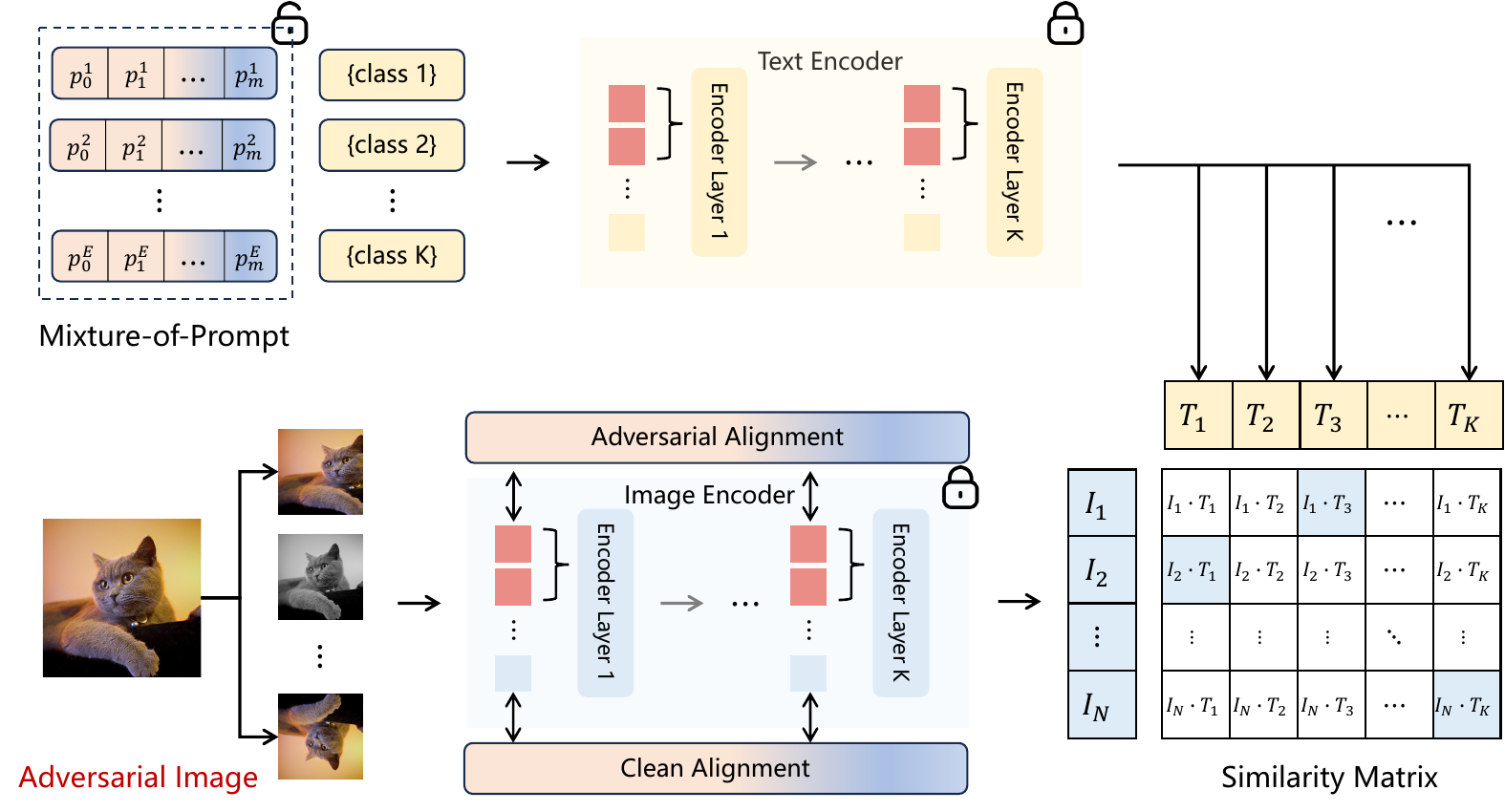}
    \caption{An overview of our proposed TAME method. Given an adversarial image, TAME generates multiple augmented views of the image and retains only those views with low entropy in their averaged prediction probabilities. During inference, TAME then optimizes the prompt by minimizing multi-view entropy across these selected views while aligning their embedding distribution with pre-computed adversarial-clean statistics from a public dataset (ImageNet).}
    \label{fig3}
\end{figure*}

\subsection{Adversarial Prompt Learning}
Adversarial Prompt Tuning (APT) integrates adversarial training into prompt tuning to enhance the adversarial robustness of CLIP on downstream tasks. Instead of relying on hand-crafted templates such as $\texttt{"a photo of a <class>"}$, APT learns robust prompts directly from the training data, thereby improving adversarial robustness on the target task. As illustrated in Figure~\ref{fig2}, APT admits four representative single-prompt designs, which differ in where the learnable prompts are inserted: (a) \emph{Textual Prompt} (text branch only), (b) \emph{Visual Prompt} (vision branch only), (c) \emph{V-L Joint Prompt} (a coupled prompt shared across both branches), and (d) \emph{V-L Independent Prompt} (separate prompts for each branch). Building upon these single-prompt variants, we further introduce the Multimodal Mixture of Prompt (MMoP) formulation, which yields four corresponding extensions: (e) \emph{Mixture of Textual Prompt}, (f) \emph{Mixture of Visual Prompt}, (g) \emph{Mixture of V-L Joint Prompt}, and (h) \emph{Mixture of V-L Independent Prompt}. In each MMoP variant, a bank of expert prompts is dynamically aggregated on a per-input basis, affording substantially greater expressive capacity than their single-prompt counterparts.

Formally, APT optimizes the prompt set $\bm{P} = \{\bm{P_v}, \bm{P_t}\}$, where $\bm{P_v}\in\mathbb{R}^{C_v \times D}$ and $\bm{P_t}\in\mathbb{R}^{C_t \times D}$ denote the visual and textual prompts, respectively, $C_v$ and $C_t$ are the corresponding numbers of prompt tokens, and $D$ is the embedding dimension. Given a downstream training set $\mathcal{D}_{\text{train}} = \{(\rvx,y)\}$, the learnable visual prompt $\bm{P_v}$ is appended to the visual input tokens to form the sequence $\{\mathbf{Z}_v(\rvx), \bm{P_v}\} = \{\textsc{cls}, e_1, e_2, \cdots, e_{T_v}, \bm{P_v}\}$, and the textual prompt $\bm{P_t}$ is analogously appended to the textual input to form $\{y, \bm{P_t}\}$. APT promotes adversarial robustness by aligning clean text embeddings with adversarial image embeddings through a min-max optimization:
\begin{equation}
\label{eq:apt}
   \argmin_{\bm{P}} \mathbb{E}_{(\rvx,y)\sim {\mathcal{D}_{\text{train}}}} \max_{\|\rvx' - \rvx\|_{\infty} \leq \epsilon} \Ls(\mathcal{I}(\rvx', \bm{P_v}), \mathcal{T}(y, \bm{P_t})),
\end{equation}
where $\mathcal{I}(\rvx', \bm{P_v})$ and $\mathcal{T}(y, \bm{P_t})$ denote the adversarial image and text embeddings, respectively. APT methods tune the prompts on the training set of each downstream task; at test time, the learned prompts are frozen and reused across all test samples.

\section{TAME}
\label{sec:method}
\subsection{Framework Overview}
Figure~\ref{fig3} illustrates the proposed \textbf{TAME} framework, which robustifies VLMs on-the-fly through an input-conditioned multimodal mixture-of-prompts architecture. Unlike train-time prompt defenses that learn a static prompt from labeled downstream data, TAME requires no downstream training samples or annotations. Instead, it relies only on the unlabeled test image and pre-computed adversarial and clean statistics from a public dataset (e.g., ImageNet). Given a test image $\rvx\in\mathcal{D}_{\text{test}}$, TAME first generates $M$ stochastic augmented views $\{\mathcal{A}_1(\rvx), \mathcal{A}_2(\rvx), \dots, \mathcal{A}_M(\rvx)\}$ and then selects the $\tau$ lowest-entropy views to form a reliable subset 
$\mathcal{H}_{\tau}(\rvx)\subseteq\{\mathcal{A}_{j}(\rvx)\}_{j=1}^{M}$. Based on these views, TAME optimizes the mixture-of-prompts $\widetilde{\bm{P}}(\rvx)$ by minimizing three objectives: (1) multi-view entropy minimization, (2) adversarial-clean alignment, and (3) MoE regularization. The adapted prompts are used only for the current test sample or batch and are reset before processing the next one. The overall procedure is summarized in Algorithm~\ref{algorithm}.

\subsection{Multimodal Mixture of Prompts}
A key limitation of single-prompt defenses is their restricted expressive capacity. A single static prompt is expected to encode a universal defense strategy for all possible inputs, which is often insufficient against adversarial perturbations that are highly diverse and input-dependent. To address this, TAME introduces a Multimodal Mixture of Prompts (MMoP), which maintains multiple expert prompts and dynamically composes an input-conditioned prompt for each test image.

\subsubsection{Expert Prompt Set}
MMoP maintains a bank of $E$ expert prompts$\{\bm{P}^{e}\}_{e=1}^{E} = \{ \bm{P}^{1}, \bm{P}^{2}, \ldots, \bm{P}^{E} \}$, each comprising a visual and a textual component $\bm{P}^{e}=(\bm{P}_{v}^{e},\bm{P}_{t}^{e})$. Depending on the modality being prompted, MMoP admits three variants:
\begin{equation}
    \{\bm{P}_{v}^{e}, \bm{P}_{t}^{e}\}_{e=1}^{E} =
    \begin{cases}
     \{\bm{P}_{v}^{e}, \varnothing\}_{e=1}^{E}, 
    & \text{(i) V}, \\[4pt]
    \{f(\bm{P}_{t}^{e}), \bm{P}_{t}^{e}\}_{e=1}^{E}, 
    & \text{(ii) VLJ}, \\[4pt]
    \{\bm{P}_{v}^{e}, \bm{P}_{t}^{e}\}_{e=1}^{E}, 
    & \text{(iii) VLI}.
    \end{cases}
    \label{eq:MMoP_designs}
\end{equation}
where $\varnothing$ indicates that no textual prompt tokens are inserted, and $f(\cdot)$ is a learnable projection that maps textual prompt tokens to the visual embedding space.

The expert bank can be initialized in several ways, including
(i) random initialization, 
(ii) seeding from hand-crafted templates (e.g., $\texttt{"a photo of a <class>"}$), or 
(iii) warm-start initialization from robust prompt defenses trained on a public dataset. 
In practice, we adopt warm-start initialization, which injects robustness priors before adaptation. Importantly, MMoP is agnostic to the underlying prompt design and recovers single-prompt defenses when $E=1$.

\subsubsection{Routing Network}
To dynamically combine expert prompts for each test image, TAME employs a lightweight routing network $g_{\phi}$. The router predicts sample-specific mixture weights from the frozen backbone tokens $\mathbf{Z}(\rvx)\in\mathbb{R}^{T\times D}$ (excluding learnable prompt tokens):
\begin{align}
    \mathbf{G}(\rvx)
    &= g_{\phi}\big(\mathbf{Z}(\rvx)\big)
    \in \mathbb{R}^{T \times E}, \label{eq:gate_token_G} \\
    \bm{\pi}(\rvx)
    &= \frac{1}{T}\sum_{t=1}^{T}
       \softmax\big(\mathbf{G}_t(\rvx)\big), \label{eq:gate_token_pi}
\end{align}
where $\mathbf{G}_{t}(\rvx)$ denotes the expert logits for token $t$, and $\bm{\pi}(\rvx)\in\mathbb{R}^{E}$ is the resulting mixture distribution. Token-wise routing allows the gate to aggregate fine-grained spatial or semantic evidence before assigning expert weights. The mixed prompts are then assembled as a convex combination of experts:
\begin{equation}
\label{eq:mixed_prompt}
    \widetilde{\bm{P}}_{v}(\rvx)
    =
    \sum_{e=1}^{E}\pi_e(\rvx)\bm{P}_{v}^{e},
    \qquad
    \widetilde{\bm{P}}_{t}(\rvx)
    =
    \sum_{e=1}^{E}\pi_e(\rvx)\bm{P}_{t}^{e}.
\end{equation}
inserted into their respective token sequences and processed by the subsequent self-attention and MLP sub-layers in our residual MMoP block. During test-time optimization, only the prompt bank $\{\bm{P}^{e}\}_{e=1}^{E}$ and the routing network $g_{\phi}$ are updated, while the backbone remains frozen, allowing TAME to adaptively select or combine complementary robustness behaviors per sample.

This mixture formulation provides a mechanistic advantage over single-prompt adaptation. A single prompt defines only one correction direction in the prompt-conditioned feature space, which may be insufficient for adversarial shifts that vary across samples, categories, and attack patterns. In contrast, TAME represents the adapted prompt as a convex combination of expert prompts, enabling the router to dynamically combine complementary robustness behaviors according to the current tokens. Together with multi-view entropy consistency, adversarial-clean alignment, and MoE regularization, this structured mixture space supports more stable test-time robustness under diverse adversarial perturbations.

\subsection{Test-Time Robustness}
\subsubsection{Multi-View Entropy-Based Sample Selection}
Following our prior works~\cite{wang2025tapt}, TAME first filters out unreliable augmented views $\mathcal{A}(\rvx)$ (e.g., when a random crop removes essential image content) via a confidence-based selection rule parameterized by a threshold $\tau$. This filter retains only augmented views with low prediction entropy (high-confidence predictions). Concretely, only views whose prediction entropy falls below the top-$\tau$ quantile (i.e., the most confident predictions) are retained $\mathcal{H}_\tau(\rvx) = \big\{\mathcal{A}_j(\rvx) \,\big|\, 1 \le j \le M,\; H(\mathcal{A}_j(\rvx)) \le H_\tau\big\}$, where $H_\tau$ denotes the entropy value corresponding to the top-$\tau$ lowest entropies among the $M$ augmentations. TAME then optimizes the prompts by minimizing the entropy of the average prediction across the selected views:
\begin{equation}
\label{eq:entropy}
    \Ls_{\text{entropy}} = -\sum_{i=1}^{K} \tilde{p}(y_i \mid \mathcal{H}_\tau(\rvx),\widetilde{\bm{P}}) \log \tilde{p}(y_i \mid \mathcal{H}_\tau(\rvx),\widetilde{\bm{P}}),
\end{equation}
where $K$ is the number of classes and $\tilde{p}(y_i \mid \mathcal{H}_\tau(\rvx), \widetilde{\bm{P}}) = \tfrac{1}{|\mathcal{H}_\tau(\rvx)|} \sum_{\hat{\rvx} \in \mathcal{H}_\tau(\rvx)} p(y_i \mid \hat{\rvx}, \widetilde{\bm{P}})$
denotes the mean predicted probability for class $y_i$ across the selected views under prompt $\widetilde{\bm{P}}$. This objective encourages cross-view prediction consistency by adjusting the prompt to each individual test instance.

\begin{table*}[ht!]
  \centering
  \caption{Zero-shot adversarial robustness (\%) of CLIP ViT-B/16 across 11 datasets under PGD, CW, DI, and AutoAttack. TAPT, MMoP, and TAME are evaluated under the same V-L Independent configuration.}
  \label{tab:vitb16_final}
  \small
  \setlength{\tabcolsep}{4pt}
  \renewcommand{\arraystretch}{1.08}
  \begin{adjustbox}{max width=\textwidth}
  \begin{tabular}{@{} ll *{12}{c} @{}}
    \toprule
    \multirow{2}{*}{\textbf{Attack}} & \multirow{2}{*}{\textbf{Method}}
    & \multicolumn{12}{c}{\textbf{Dataset}} \\
    \cmidrule(lr){3-14}
    & 
    & \makecell[c]{ImageNet}
    & \makecell[c]{Caltech101}
    & \makecell[c]{DTD}
    & \makecell[c]{EuroSAT}
    & \makecell[c]{Pets}
    & \makecell[c]{Aircraft}
    & \makecell[c]{Food101}
    & \makecell[c]{Flowers}
    & \makecell[c]{Cars}
    & \makecell[c]{SUN397}
    & \makecell[c]{UCF101}
    & \makecell[c]{Avg.} \\
    \midrule

    \multirow{11}{*}{\textbf{PGD}}
      & Vanilla      & 1.4 & 21.3 & 1.4 & 6.0 & 5.0 & 0.0 & 9.8 & 1.6 & 0.7 & 0.8 & 1.8 & 4.5 \\
      & APT-T        & 1.5 & 14.2 & 2.1 & 0.0 & 1.5 & 0.0 & 0.3 & 0.3 & 0.2 & 0.9 & 0.5 & 2.0 \\
      & APT-V        & 19.4 & 61.2 & 18.5 & 8.0 & 37.4 & 3.9 & 12.1 & 25.1 & 9.8 & 17.2 & 16.4 & 20.8 \\
      & APT-VLJ      & 23.9 & 61.7 & 18.7 & 9.7 & 41.4 & 3.2 & 14.1 & 23.4 & 12.2 & 17.7 & 15.5 & 22.0 \\
      & APT-VLI      & 24.3 & 65.3 & 18.9 & 10.0 & 43.6 & 3.1 & 14.3 & 23.6 & 10.5 & 18.2 & 17.4 & 22.7 \\
      & FAP          & 23.2 & 63.0 & 11.2 & 10.5 & 43.7 & 4.2 & 14.2 & 23.2 & 10.7 & 19.2 & 20.4 & 22.1 \\
      & NAP-Tuning   & 27.2 & 52.9 & 10.8 & 4.4 & 43.0 & 1.3 & 9.1 & 6.1 & 4.5 & 11.4 & 14.1 & 16.8 \\
      & R-TPT        & 54.8 & 78.6 & 35.4 & 26.0 & 74.4 & 12.2 & 62.9 & 51.6 & 42.9 & 49.1 & 50.3 & 48.9 \\
      & MMoP         & 28.0 & 67.1 & 20.2 & 11.3 & 48.3 & 3.8 & 22.2 & 24.6 & 11.7 & 22.6 & 23.1 & 25.7 \\
      & TAPT         & 50.0 & 79.0 & 32.4 & \textbf{36.2} & 67.5 & \textbf{13.1} & 65.7 & 49.8 & 39.6 & 48.3 & 47.5 & 48.1 \\
      & \textbf{TAME} & \textbf{61.7} & \textbf{84.3} & \textbf{36.3} & 26.5 & \textbf{78.4} & 12.5 & \textbf{66.3} & \textbf{52.3} & \textbf{45.3} & \textbf{50.4} & \textbf{53.1} & \textbf{51.6} \\
    \midrule

    \multirow{11}{*}{\textbf{CW}}
      & Vanilla      & 0.2 & 0.4 & 0.2 & 0.0 & 0.0 & 0.0 & 0.0 & 0.0 & 0.0 & 0.0 & 0.0 & 0.1 \\
      & APT-T        & 0.3 & 0.3 & 0.6 & 0.0 & 0.0 & 0.3 & 0.0 & 0.1 & 0.0 & 0.2 & 0.0 & 0.2 \\
      & APT-V        & 4.9 & 22.3 & 12.1 & 1.6 & 4.0 & 1.8 & 2.3 & 9.3 & 2.1 & 7.2 & 2.3 & 6.4 \\
      & APT-VLJ      & 5.5 & 21.2 & 10.6 & 9.2 & 4.3 & 1.1 & 2.3 & 7.1 & 5.7 & 7.6 & 3.6 & 7.1 \\
      & APT-VLI      & 7.4 & 23.2 & 16.3 & 2.3 & 5.8 & 2.5 & 3.1 & 9.9 & 7.4 & 10.0 & 5.0 & 8.4 \\
      & FAP          & 10.4 & 32.2 & 8.9 & 9.4 & 17.0 & 2.0 & 5.6 & 15.7 & 7.1 & 12.6 & 7.9 & 11.7 \\
      & NAP-Tuning   & 3.2 & 12.5 & 5.4 & 3.6 & 1.3 & 0.7 & 1.6 & 2.4 & 2.3 & 3.2 & 1.0 & 3.4 \\
      & R-TPT        & 54.6 & 78.0 & 30.0 & 26.3 & \textbf{74.5} & 20.7 & 63.9 & \textbf{56.7} & 50.9 & 54.3 & 56.3 & 51.5 \\
      & MMoP         & 12.9 & 19.2 & 14.5 & 9.9 & 5.0 & 2.0 & 5.4 & 13.5 & 7.3 & 14.5 & 12.8 & 10.6 \\
      & TAPT         & 58.9 & 80.5 & 35.6 & \textbf{35.4} & 67.4 & 20.0 & \textbf{72.5} & 56.4 & 54.3 & 56.4 & 58.2 & 54.1 \\
      & \textbf{TAME} & \textbf{60.7} & \textbf{84.7} & \textbf{37.1} & 29.9 & 72.9 & \textbf{23.0} & 65.5 & 56.5 & \textbf{55.7} & \textbf{59.7} & \textbf{59.4} & \textbf{55.0} \\
    \midrule

    \multirow{11}{*}{\textbf{DI}}
      & Vanilla      & 6.1 & 25.8 & 8.7 & 0.3 & 11.1 & 0.5 & 9.5 & 3.0 & 3.4 & 4.9 & 4.7 & 7.1 \\
      & APT-T        & 10.0 & 38.2 & 8.0 & 0.4 & 10.6 & 1.6 & 6.7 & 5.2 & 4.9 & 7.6 & 5.4 & 9.0 \\
      & APT-V        & 29.2 & 67.5 & 23.5 & 12.3 & 47.1 & 5.0 & 21.8 & 30.2 & 18.1 & 27.1 & 22.0 & 27.6 \\
      & APT-VLJ      & 34.9 & 72.0 & 25.0 & 10.5 & 52.8 & 4.2 & 24.6 & 32.9 & 18.9 & 28.5 & 24.8 & 29.9 \\
      & APT-VLI      & 35.1 & 67.7 & 19.3 & 10.6 & 49.8 & 4.3 & 23.6 & 27.7 & 16.2 & 26.4 & 21.5 & 27.5 \\
      & FAP          & 32.3 & 73.4 & 15.2 & 11.1 & 54.5 & 8.3 & 22.9 & 31.9 & 17.3 & 27.6 & 28.9 & 29.4 \\
      & NAP-Tuning   & 28.8 & 55.7 & 12.4 & 5.2 & 46.8 & 1.7 & 10.5 & 7.0 & 5.5 & 13.2 & 16.1 & 18.4 \\
      & R-TPT        & 46.9 & 82.5 & 31.5 & 14.9 & 67.1 & 11.7 & 55.2 & 49.6 & 40.8 & 47.5 & 48.5 & 45.1 \\
      & MMoP         & 39.7 & 74.0 & 26.5 & 13.8 & 61.5 & 5.9 & 32.8 & 37.0 & 23.3 & 33.7 & 33.7 & 34.7 \\
      & TAPT         & 53.8 & 80.5 & 32.3 & \textbf{39.6} & 69.4 & 13.3 & \textbf{70.6} & 51.1 & 42.5 & \textbf{50.3} & 48.2 & 50.1 \\
      & \textbf{TAME} & \textbf{59.4} & \textbf{83.0} & \textbf{35.2} & 29.0 & \textbf{75.1} & \textbf{13.6} & 70.5 & \textbf{53.0} & \textbf{43.5} & 48.4 & \textbf{49.2} & \textbf{50.9} \\
    \midrule

    \multirow{11}{*}{\makecell[c]{\textbf{AA}}}
    & Vanilla      & 0.0 & 0.0 & 0.0 & 0.1 & 0.0 & 0.1 & 0.0 & 0.0 & 0.0 & 0.0 & 0.0 & 0.1 \\
      & APT-T        & 0.0 & 0.0 & 0.1 & 0.0 & 0.0 & 0.0 & 0.0 & 0.0 & 0.0 & 0.0 & 0.0 & 0.0 \\
      & APT-V        & 14.8 & 55.9 & 15.2 & 2.3 & 31.3 & 2.3 & 8.3 & 18.0 & 5.8 & 12.4 & 12.7 & 16.3 \\
      & APT-VLJ      & 16.5 & 53.2 & 12.6 & 5.6 & 31.3 & 1.7 & 8.0 & 16.0 & 4.9 & 11.1 & 11.1 & 15.6 \\
      & APT-VLI      & 17.2 & 57.1 & 14.4 & 8.2 & 35.3 & 1.5 & 8.9 & 16.6 & 5.1 & 11.9 & 12.5 & 17.2 \\
      & FAP          & 15.8 & 57.5 & 9.0 & 6.1 & 33.6 & 1.8 & 8.7 & 15.8 & 4.2 & 12.3 & 13.6 & 16.2 \\
      & NAP-Tuning   & 24.9 & 49.5 & 9.9 & 2.7 & 40.0 & 0.7 & 7.9 & 5.5 & 3.3 & 9.9 & 11.5 & 15.1 \\
      & R-TPT        & 50.0 & 80.5 & 31.1 & 30.2 & 77.9 & \textbf{20.9} & 71.9 & 49.7 & 54.2 & 48.0 & 51.5 & 51.4 \\
      & MMoP         & 10.8 & 26.9 & 12.4 & 10.0 & 5.9 & 1.8 & 7.5 & 14.5 & 5.5 & 12.8 & 11.0 & 10.8 \\
      & TAPT         & 55.1 & 80.3 & 35.7 & \textbf{44.4} & 70.2 & 16.0 & \textbf{76.2} & \textbf{55.1} & 50.5 & \textbf{53.6} & 54.5 & 53.8 \\
      & \textbf{TAME} & \textbf{60.8} & \textbf{81.2} & \textbf{36.8} & 36.8 & \textbf{78.5} & 18.7 & 71.5 & 50.3 & \textbf{56.0} & 49.5 & \textbf{55.0} & \textbf{54.1} \\
    \bottomrule
  \end{tabular}
  \end{adjustbox}
  \vspace{-13pt}
\end{table*}

\begin{table*}[ht!]
  \centering
  \caption{Zero-shot adversarial robustness (\%) of CLIP ViT-B/32 across 11 datasets under PGD, CW, DI, and AutoAttack. TAPT, MMoP, and TAME are evaluated under the same V-L Independent configuration.}
  \label{tab:vitb32_final}
  \small
  \setlength{\tabcolsep}{4pt}
  \renewcommand{\arraystretch}{1.08}
  \begin{adjustbox}{max width=\textwidth}
  \begin{tabular}{@{} ll *{12}{c} @{}}
    \toprule
    \multirow{2}{*}{\textbf{Attack}} & \multirow{2}{*}{\textbf{Method}}
    & \multicolumn{12}{c}{\textbf{Dataset}} \\
    \cmidrule(lr){3-14}
    & 
    & \makecell[c]{ImageNet}
    & \makecell[c]{Caltech101}
    & \makecell[c]{DTD}
    & \makecell[c]{EuroSAT}
    & \makecell[c]{Pets}
    & \makecell[c]{Aircraft}
    & \makecell[c]{Food101}
    & \makecell[c]{Flowers}
    & \makecell[c]{Cars}
    & \makecell[c]{SUN397}
    & \makecell[c]{UCF101}
    & \makecell[c]{Avg.} \\
    \midrule

    \multirow{11}{*}{\textbf{PGD}}
      & Vanilla      & 1.3 & 22.9 & 5.0 & 0.0 & 2.6 & 0.0 & 3.6 & 1.5 & 0.2 & 1.1 & 1.7 & 3.6 \\
      & APT-T        & 2.8 & 22.7 & 3.9 & 0.0 & 3.0 & 0.1 & 0.9 & 3.2 & 0.7 & 1.8 & 2.0 & 3.7 \\
      & APT-V        & 18.9 & 63.8 & 20.2 & 0.6 & 36.1 & 2.7 & 15.0 & 23.1 & 9.4 & 19.1 & 18.4 & 20.7 \\
      & APT-VLJ      & 21.4 & 64.3 & 14.7 & 10.4 & 37.7 & 2.0 & 16.5 & 21.0 & 8.9 & 17.5 & 18.0 & 21.1 \\
      & APT-VLI      & 21.2 & 63.2 & 15.8 & 10.4 & 37.6 & 1.5 & 16.1 & 20.2 & 8.3 & 16.9 & 17.6 & 20.8 \\
      & FAP          & 23.4 & 63.5 & 15.8 & 7.2 & 43.1 & 2.3 & 18.4 & 27.9 & 12.7 & 22.0 & 23.7 & 23.6 \\
      & NAP-Tuning   & 21.6 & 49.5 & 8.9 & 7.7 & 31.5 & 0.7 & 7.7 & 7.7 & 2.5 & 9.9 & 11.8 & 14.5 \\
      & R-TPT        & 52.1 & 75.9 & 26.4 & 20.0 & 73.4 & 4.3 & 61.8 & 42.9 & 40.8 & 43.8 & 44.3 & 44.2 \\
      & MMoP         & 25.3 & 63.5 & 15.8 & 7.8 & 43.6 & 2.9 & 21.9 & 23.9 & 12.7 & 22.4 & 22.0 & 23.8 \\
      & TAPT         & 48.2 & 82.1 & 30.7 & \textbf{31.6} & 68.1 & 4.5 & \textbf{64.7} & 44.6 & 41.3 & 47.6 & 49.1 & 46.6 \\
      & \textbf{TAME} & \textbf{57.0} & \textbf{82.9} & \textbf{31.1} & 22.2 & \textbf{76.0} & \textbf{8.4} & 60.8 & \textbf{47.3} & \textbf{42.2} & \textbf{48.9} & \textbf{49.9} & \textbf{47.9} \\
    \midrule

    \multirow{11}{*}{\textbf{CW}}
      & Vanilla      & 0.3 & 1.2 & 0.9 & 0.0 & 0.0 & 0.0 & 0.0 & 0.2 & 0.0 & 0.2 & 0.4 & 0.3 \\
      & APT-T        & 0.9 & 3.4 & 1.7 & 0.0 & 0.5 & 0.1 & 0.1 & 1.0 & 0.2 & 1.4 & 1.1 & 0.9 \\
      & APT-V        & 6.0 & 27.7 & 15.7 & 0.9 & 6.8 & 2.5 & 2.3 & 7.8 & 2.3 & 9.7 & 3.0 & 7.7 \\
      & APT-VLJ      & 6.8 & 25.0 & 12.1 & 11.5 & 7.2 & 2.4 & 3.6 & 7.1 & 7.2 & 8.9 & 3.9 & 8.7 \\
      & APT-VLI      & 6.7 & 22.8 & 12.3 & 11.5 & 9.2 & 1.5 & 3.0 & 6.6 & 5.8 & 9.0 & 4.5 & 8.4 \\
      & FAP          & 14.4 & 43.9 & 12.8 & 8.6 & 25.2 & 1.7 & 11.2 & 23.3 & 9.8 & 17.0 & 13.7 & 16.5 \\
      & NAP-Tuning   & 4.8 & 17.0 & 6.4 & 2.9 & 1.7 & 0.3 & 1.5 & 3.8 & 2.0 & 4.2 & 1.8 & 4.2 \\
      & R-TPT        & 54.1 & 75.5 & 27.0 & 25.9 & 74.8 & 6.3 & 65.3 & 45.2 & 40.0 & 45.7 & 46.1 & 46.0 \\
      & MMoP         & 9.5 & 16.7 & 10.2 & 6.5 & 7.1 & 1.6 & 4.9 & 6.8 & 5.6 & 12.2 & 10.8 & 8.4 \\
      & TAPT         & 52.9 & 82.1 & 33.5 & \textbf{30.6} & 70.5 & 6.9 & \textbf{74.1} & 49.2 & 49.0 & \textbf{54.4} & 46.2 & 49.9 \\
      & \textbf{TAME} & \textbf{58.1} & \textbf{82.2} & \textbf{34.0} & 20.6 & \textbf{77.1} & \textbf{9.5} & 69.6 & \textbf{49.7} & \textbf{49.5} & 50.2 & \textbf{49.4} & \textbf{50.0} \\
    \midrule

    \multirow{11}{*}{\textbf{DI}}
      & Vanilla      & 6.4 & 37.2 & 11.1 & 0.7 & 9.9 & 0.1 & 7.5 & 9.3 & 3.9 & 8.6 & 5.8 & 9.1 \\
      & APT-T        & 10.5 & 51.8 & 11.5 & 0.8 & 13.1 & 0.7 & 8.9 & 16.2 & 6.4 & 9.8 & 9.1 & 12.6 \\
      & APT-V        & 26.9 & 68.6 & 22.0 & 5.1 & 46.0 & 4.8 & 23.7 & 26.5 & 13.9 & 26.4 & 24.6 & 26.2 \\
      & APT-VLJ      & 30.4 & 69.5 & 17.0 & 11.9 & 47.9 & 2.9 & 26.7 & 26.1 & 17.3 & 26.5 & 26.4 & 27.5 \\
      & APT-VLI      & 28.9 & 69.5 & 20.3 & 10.9 & 47.0 & 2.8 & 24.8 & 24.3 & 14.8 & 23.8 & 23.6 & 26.4 \\
      & FAP          & 29.0 & 69.0 & 18.6 & 9.2 & 48.7 & 3.2 & 23.9 & 32.9 & 17.3 & 27.8 & 27.4 & 27.9 \\
      & NAP-Tuning   & 25.5 & 52.0 & 10.2 & 8.0 & 34.1 & 1.0 & 8.9 & 8.9 & 3.1 & 11.3 & 13.5 & 16.0 \\
      & R-TPT        & 49.6 & 83.2 & 34.5 & 12.4 & 69.6 & 6.4 & 59.4 & 41.3 & 41.5 & \textbf{50.6} & 50.1 & 45.3 \\
      & MMoP         & 35.7 & 70.8 & 21.0 & 11.8 & 57.9 & 5.9 & 30.9 & 32.5 & 20.7 & 30.5 & 32.3 & 31.8 \\
      & TAPT         & 50.0 & 82.7 & 31.5 & \textbf{32.7} & 69.4 & 4.4 & \textbf{65.6} & 45.1 & 42.8 & 49.1 & 49.9 & 47.6 \\
      & \textbf{TAME} & \textbf{56.0} & \textbf{83.7} & \textbf{35.1} & 28.9 & \textbf{73.9} & \textbf{8.3} & 58.1 & \textbf{46.0} & \textbf{44.5} & 49.6 & \textbf{50.6} & \textbf{48.6} \\
    \midrule

    \multirow{11}{*}{\makecell[c]{\textbf{AA}}}
      & Vanilla      & 0.0 & 0.0 & 0.0 & 0.1 & 0.0 & 0.1 & 0.0 & 0.1 & 0.1 & 0.0 & 0.1 & 0.0 \\
      & APT-T        & 0.1 & 0.4 & 0.3 & 0.0 & 0.1 & 0.0 & 0.0 & 0.0 & 0.0 & 0.1 & 0.1 & 0.1 \\
      & APT-V        & 8.3 & 44.9 & 12.7 & 0.1 & 16.4 & 0.5 & 5.5 & 9.0 & 2.7 & 7.5 & 7.8 & 10.5 \\
      & APT-VLJ      & 10.3 & 46.2 & 10.0 & 3.0 & 18.5 & 0.6 & 6.7 & 9.5 & 1.7 & 7.4 & 7.2 & 11.0 \\
      & APT-VLI      & 9.7 & 45.6 & 10.6 & 6.8 & 17.3 & 0.4 & 6.0 & 8.7 & 2.0 & 6.8 & 6.9 & 11.0 \\
      & FAP          & 17.3 & 60.3 & 14.0 & 4.0 & 35.7 & 1.0 & 12.3 & 23.0 & 6.3 & 16.1 & 18.3 & 18.9 \\
      & NAP-Tuning   & 19.6 & 48.0 & 8.4 & 2.0 & 29.0 & 0.5 & 7.1 & 6.7 & 1.9 & 8.9 & 10.8 & 13.0 \\
      & R-TPT        & 46.1 & 76.5 & 28.6 & 26.1 & 66.8 & 6.2 & 68.1 & 45.5 & \textbf{50.8} & 46.8 & 47.8 & 46.3 \\
      & MMoP         & 8.7 & 31.0 & 3.0 & 1.1 & 9.9 & 0.1 & 1.5 & 1.9 & 0.1 & 2.8 & 3.6 & 5.8 \\
      & TAPT         & 49.7 & \textbf{80.6} & \textbf{33.3} & \textbf{38.1} & 68.4 & 5.0 & 67.5 & 47.1 & 48.1 & \textbf{50.3} & 50.4 & 49.0 \\
      & \textbf{TAME} & \textbf{54.0} & 80.3 & 30.1 & 31.5 & \textbf{71.3} & \textbf{8.8} & \textbf{69.7} & \textbf{47.2} & 50.6 & 45.8 & \textbf{50.9} & \textbf{49.1} \\
    \bottomrule
  \end{tabular}
  \end{adjustbox}
  \vspace{-10pt}
\end{table*}

\subsubsection{Adversarial-Clean Alignment}
Adversarial perturbations can substantially shift the embeddings $\mathcal{I}(\rvx, \widetilde{\bm{P}})$ produced by the image encoder away from those of clean inputs, thereby misleading downstream prediction. To counter this distributional drift, we align the first- and second-order statistics of the test-time embeddings with pre-computed reference statistics derived from a public dataset $\mathcal{D}_{\text{public}}$. Although the ideal reference would be CLIP's original pre-training data, this data is not publicly available; we therefore adopt ImageNet as a proxy due to its large scale and broad semantic coverage~\cite{bahng2022exploring}. The per-layer mean and variance of the current embeddings are computed as:
\begin{align}
\label{eq:dist}
    \mu_l(\mathcal{H}_\tau(\rvx);\widetilde{\bm{P}}) &= \frac{1}{|\mathcal{H}_\tau(\rvx)|} \sum_{\hat{\rvx} \in \mathcal{H}_\tau(\rvx)} \mathcal{I}_l(\hat{\rvx},\widetilde{\bm{P}}), \\
    \sigma^2_l(\mathcal{H}_\tau(\rvx);\widetilde{\bm{P}}) &= \frac{\sum_{\hat{\rvx} \in \mathcal{H}_\tau(\rvx)} (\mathcal{I}_l(\hat{\rvx},\widetilde{\bm{P}}) - \mu_l(\mathcal{H}_\tau(\rvx);\widetilde{\bm{P}}))^2}{|\mathcal{H}_\tau(\rvx)| - 1}
\end{align}
where $\mathcal{I}_l(\hat{\rvx},\widetilde{\bm{P}})$ denotes the layer-$l$ embedding of view $\hat{\rvx}$ under prompt $\widetilde{\bm{P}}$. Analogously, we pre-compute the layer-wise statistics on $\mathcal{D}_{\text{public}}$ using the robust prompt $\widetilde{\bm{P}}_{\text{adv}}$ (obtained via APT on the public data) and the clean prompt $\widetilde{\bm{P}}_{\text{clean}}$ (obtained via standard prompt tuning on the public data), yielding the offline references $\{\mu_{\text{adv}}, \sigma^2_{\text{adv}}\}$ and $\{\mu_{\text{clean}}, \sigma^2_{\text{clean}}\}$. The alignment objectives are then defined as:
\begin{align}
\label{eq:align}
    \Ls_{\text{adv}} &= \frac{1}{L} \sum_{l=1}^{L} \big(\lVert \mu_l - \mu_{\text{adv},l}\rVert_1 + \lVert \sigma^2_l - \sigma^2_{\text{adv},l} \rVert_1\big),   \\
    \Ls_{\text{clean}} &= \frac{1}{L} \sum_{l=1}^{L} \big(\lVert \mu_l - \mu_{\text{clean},l} \rVert_1 + \lVert \sigma^2_l - \sigma^2_{\text{clean},l} \rVert_1\big),   \\
    \Ls_{\text{align}} &= \alpha,\Ls_{\text{adv}} + (1-\alpha),\Ls_{\text{clean}},
\end{align}
where $\alpha \in [0,1]$ balances robustness and clean accuracy. Setting $\alpha=0$ favors the clean distribution at the cost of robustness, whereas $\alpha=1$ prioritizes adversarial alignment.

\subsubsection{MoE Regularization}
\label{sec:moe_reg}
Without regularization~\cite{shazeer2017outrageously,fedus2022switch,jacobs1991adaptive}, the soft-routing TAME may suffer from router collapse, where $\bm{\pi}(\rvx)$ concentrates on a single expert, and expert redundancy, where $\{\bm{P}^{e}\}_{e=1}^{E}$ degenerates into near-duplicate prompts. Unlike conventional batch-level balancing in train-time MoE, the test-time setting offers no mini-batch over which to spread routing mass; we therefore propose an instance-level balance loss computed over the multi-view set $\mathcal{H}_\tau(\rvx)$ of the current sample, paired with a prompt diversity loss on the expert bank:
\begin{equation}
\label{eq:moe_losses}
    \Ls_{\text{bal}} = \tfrac{1}{E}\!\sum_{e=1}^{E}\!\Big(\bar{\pi}_e - \tfrac{1}{E}\Big)^{2}\!,\quad
    \Ls_{\text{div}} = \tfrac{2}{E(E\!-\!1)}\!\sum_{i<j}\!\big[\cos(\bm{p}^{i},\bm{p}^{j})\big]_+,
\end{equation}
where $\bar{\bm{\pi}}=\tfrac{1}{|\mathcal{H}_\tau(\rvx)|}\sum_{\hat{\rvx}}\bm{\pi}(\hat{\rvx})$ is the view-averaged routing distribution and $[\cdot]_+$ penalizes only positively correlated pairs so that anti-correlated experts remain free to encode complementary robustness behaviors. Aggregating across all TAME blocks $\ell\!\in\!\mathcal{L}_{\text{MoE}}$ yields the final TAME objective:
\begin{equation}
\label{eq:moe_reg}
\begin{aligned}
    \Ls_{\text{MoE}} &= \gamma(t)\!\sum_{\ell\in\mathcal{L}_{\text{MoE}}}\!\big(\lambda_{\text{bal}}\Ls^{l}_{\text{bal}} + \lambda_{\text{div}}\Ls^{l}_{\text{div}}\big),\\
    \Ls_{\text{TAME}} &= \Ls_{\text{entropy}} +\Ls_{\text{align}} + \Ls_{\text{MoE}},
\end{aligned}
\end{equation}
with $\lambda_{\text{bal}},\lambda_{\text{div}}\!\ge\!0$ fixed and $\gamma(t)=\min(1,\,t/T_{\text{warm}})$ linearly ramping up the regularizer over the first $T_{\text{warm}}$ adaptation steps, preventing experts from collapsing before departing from initialization. \textit{\textbf{Note that, to ensure inference efficiency, TAME recommends only a single step of prompt tuning for each inference.}}

\section{Experiments}
\label{sec:exp}

\begin{table*}[t]
\centering
\setlength{\tabcolsep}{3.2pt}
\renewcommand{\arraystretch}{1.10}
\resizebox{\linewidth}{!}{%
\begin{tabular}{@{}cll*{12}{c}@{}}
\toprule
& \textbf{Method} &
& \textbf{ImageNet} & \textbf{Caltech} & \textbf{DTD} & \textbf{EuroSAT} & \textbf{Pets} & \textbf{Aircraft} & \textbf{Food101} & \textbf{Flowers} & \textbf{Cars} & \textbf{SUN397} & \textbf{UCF101} & \textbf{Avg.} \\
\midrule

\multirow{6}{*}{\rotatebox[origin=c]{90}{\textbf{ViT-B/16}}}
& \multirow{3}{*}{MMoP}
& V    & 22.9 & 63.0 & 18.6 & 10.6 & 43.0 & 4.3 & 12.8 & 24.9 & 10.9 & 18.9 & 17.0 & 22.4 \\
& & VLJ & 24.7 & 60.0 & 15.9 & 0.5 & 41.6 & 2.2 & 15.4 & 19.6 & 6.5 & 16.5 & 17.2 & 20.0 \\
& & VLI & 28.0 & 67.1 & 20.2 & 11.3 & 48.3 & 3.8 & 22.2 & 24.6 & 11.7 & 22.6 & 23.1 & 25.7 \\

\cmidrule(lr){2-15}
& \multirow{3}{*}{\textbf{TAME}}
& V    & 46.6\imp{23.7} & 78.2\imp{15.2} & 32.3\imp{13.7} & 26.9\imp{16.3} & 61.0\imp{18.0} & 14.0\imp{9.7} & 65.8\imp{53.0} & 46.6\imp{21.7} & 35.1\imp{24.2} & 46.6\imp{27.7} & 45.8\imp{28.8} & \textbf{45.3}\imp{22.9} \\
& & VLJ & 55.3\imp{30.6} & 74.8\imp{14.8} & 26.7\imp{10.8} & 14.1\imp{13.6} & 69.6\imp{28.0} & 8.5\imp{6.3} & 51.0\imp{35.6} & 47.8\imp{28.2} & 35.5\imp{29.0} & 37.9\imp{21.4} & 36.5\imp{19.3} & \textbf{41.6}\imp{21.6} \\
& & VLI & 61.7\imp{33.7} & 84.3\imp{17.2} & 36.3\imp{16.1} & 26.5\imp{15.2} & 78.4\imp{30.1} & 12.5\imp{8.7} & 66.3\imp{44.1} & 52.3\imp{27.7} & 45.3\imp{33.6} & 50.4\imp{27.8} & 53.1\imp{30.0} & \textbf{51.6}\imp{25.9} \\

\midrule

\multirow{6}{*}{\rotatebox[origin=c]{90}{\textbf{ViT-B/32}}}
& \multirow{3}{*}{MMoP}
& V    & 21.9 & 64.7 & 20.2 & 11.3 & 42.1 & 3.8 & 17.5 & 27.1 & 11.0 & 21.7 & 18.9 & 23.7 \\
& & VLJ & 22.6 & 61.0 & 16.1 & 1.1 & 41.0 & 2.3 & 15.1 & 18.4 & 6.7 & 17.3 & 18.1 & 20.0 \\
& & VLI & 25.3 & 63.5 & 15.8 & 7.8 & 43.6 & 2.9 & 21.9 & 23.9 & 12.7 & 22.4 & 22.0 & 23.8 \\
\cmidrule(lr){2-15}
& \multirow{3}{*}{\textbf{TAME}}
& V    & 37.1\imp{15.2} & 76.8\imp{12.1} & 28.0\imp{7.8} & 24.5\imp{13.2} & 54.6\imp{12.5} & 10.5\imp{6.7} & 47.3\imp{29.8} & 39.9\imp{12.8} & 28.0\imp{17.0} & 40.5\imp{18.8} & 41.3\imp{22.4} & \textbf{39.0}\imp{15.3} \\
& & VLJ & 46.0\imp{23.4} & 72.8\imp{11.8} & 27.4\imp{11.3} & 9.2\imp{8.1} & 69.6\imp{28.6} & 6.8\imp{4.5} & 40.1\imp{25.0} & 39.8\imp{21.4} & 23.4\imp{16.7} & 37.1\imp{19.8} & 35.6\imp{17.5} & \textbf{37.1}\imp{17.1} \\
& & VLI & 57.0\imp{31.7} & 82.9\imp{19.4} & 31.1\imp{15.3} & 22.2\imp{14.4} & 76.0\imp{32.4} & 8.4\imp{5.5} & 60.8\imp{38.9} & 47.3\imp{23.4} & 42.2\imp{29.5} & 48.9\imp{26.5} & 49.9\imp{27.9} & \textbf{47.9}\imp{24.1} \\
\midrule

\multirow{6}{*}{\rotatebox[origin=c]{90}{\textbf{ViT-L/14}}}
  & \multirow{3}{*}{MMoP}
  & V    & 36.6 & 77.4 & 24.1 & 10.4 & 63.9 & 9.0 & 27.2 & 36.6 & 26.2 & 32.3 & 33.8 & 34.3 \\
  & & VLJ & 60.6 & 88.6 & 40.6 & 19.1 & 82.0 & 13.8 & 56.3 & 54.6 & 47.3 & 54.0 & 52.7 & 51.8 \\
  & & VLI & 61.6 & 90.7 & 42.8 & 19.6 & 83.0 & 13.6 & 58.6 & 59.0 & 52.1 & 57.3 & 53.3 & 53.8 \\
  \cmidrule(lr){2-15}
  & \multirow{3}{*}{\textbf{TAME}}
  & V    & 63.5\imp{26.9} & 91.1\imp{13.7} & 43.1\imp{19.0} & 16.7\imp{6.3} & 82.2\imp{18.3} & 20.5\imp{11.5} & 61.2\imp{34.0} & 57.0\imp{20.4} & 49.7\imp{23.5} & 56.8\imp{24.5} & 60.3\imp{26.5} & \textbf{54.8}\imp{20.4} \\
  & & VLJ & 69.0\imp{8.4} & 92.2\imp{3.6} & 45.1\imp{4.5} & 20.1\imp{1.0} & 85.1\imp{3.1} & 16.6\imp{2.8} & 67.2\imp{10.9} & 55.2\imp{0.6} & 52.9\imp{5.6} & 59.9\imp{5.9} & 62.7\imp{10.0} & \textbf{56.9}\imp{5.1} \\
  & & VLI & 70.7\imp{9.1} & 93.8\imp{3.1} & 49.2\imp{6.4} & 24.0\imp{4.4} & 86.7\imp{3.7} & 16.3\imp{2.7} & 70.6\imp{12.0} & 59.4\imp{0.4} & 59.5\imp{7.4} & 62.7\imp{5.4} & 64.7\imp{11.4} & \textbf{59.8}\imp{6.0} \\

\bottomrule
\end{tabular}%
}
\caption{Zero-shot adversarial robustness (\%) under PGD attack. Green superscripts indicate the absolute improvement of TAME over MMoP baseline. V, VLJ, and VLI denote Visual Only, V-L Joint, and V-L Independent, respectively.}
\label{tab:pgd_MMoP_tame_all}
\end{table*}

\begin{table}[t]
\centering
\scriptsize
\setlength{\tabcolsep}{2.1pt}
\renewcommand{\arraystretch}{1.08}
\resizebox{\columnwidth}{!}{
\begin{tabular}{@{}lccccccccccccc@{}}
\toprule
& 
& \rot{ImageNet}
& \rot{Caltech}
& \rot{DTD}
& \rot{EuroSAT}
& \rot{Pets}
& \rot{Aircraft}
& \rot{Food101}
& \rot{Flowers}
& \rot{Cars}
& \rot{SUN397}
& \rot{UCF101}
& \rot{Avg.} \\
\midrule

\multirow{3}{*}{V}
& MMoP
& 67.3 & 93.4 & 45.1 & 41.8
& \textbf{88.3} & \textbf{24.2} & 85.7 & \textbf{67.4}
& 66.0 & 63.1 & 65.9 & 64.4 \\
& TAME
& \textbf{67.8} & \textbf{93.7} & \textbf{45.5} & \textbf{42.4}
& 88.1 & 24.0 & \textbf{86.0} & 67.2
& \textbf{66.5} & \textbf{63.5} & \textbf{66.4} & \textbf{64.6} \\
& \gain{$\Delta$}
& \gain{+0.5} & \gain{+0.3} & \gain{+0.4} & \gain{+0.6}
& \gain{-0.2} & \gain{-0.2} & \gain{+0.3} & \gain{-0.2}
& \gain{+0.5} & \gain{+0.4} & \gain{+0.5} & \gain{+0.2} \\
\midrule

\multirow{3}{*}{VLJ}
& MMoP
& 70.9 & 87.2 & 35.0 & 17.9
& \textbf{82.9} & 14.5 & 74.2 & 50.4
& 41.8 & 57.8 & 60.2 & 53.9 \\
& TAME
& \textbf{71.4} & \textbf{87.7} & \textbf{35.6} & \textbf{18.5}
& 82.7 & \textbf{15.3} & \textbf{74.8} & \textbf{51.2}
& \textbf{42.5} & \textbf{58.4} & \textbf{60.8} & \textbf{54.3} \\
& \gain{$\Delta$}
& \gain{+0.5} & \gain{+0.5} & \gain{+0.6} & \gain{+0.6}
& \gain{-0.2} & \gain{+0.8} & \gain{+0.6} & \gain{+0.8}
& \gain{+0.7} & \gain{+0.6} & \gain{+0.6} & \gain{+0.4} \\
\midrule

\multirow{3}{*}{VLI}
& MMoP
& 68.1 & 90.9 & 40.8 & 46.6
& 89.2 & \textbf{17.6} & 85.5 & \textbf{60.0}
& 64.4 & 63.1 & 67.1 & 63.0 \\
& TAME
& \textbf{68.7} & \textbf{91.3} & \textbf{41.4} & \textbf{47.2}
& \textbf{90.0} & 17.4 & \textbf{85.9} & 59.8
& \textbf{65.0} & \textbf{63.6} & \textbf{67.6} & \textbf{63.4} \\
& \gain{$\Delta$}
& \gain{+0.6} & \gain{+0.4} & \gain{+0.6} & \gain{+0.6}
& \gain{+0.8} & \gain{-0.2} & \gain{+0.4} & \gain{-0.2}
& \gain{+0.6} & \gain{+0.5} & \gain{+0.5} & \gain{+0.4} \\

\bottomrule
\end{tabular}
}
\caption{Zero-shot clean accuracy (\%) of different defense methods from ImageNet to downstream datasets, including MMoP baselines (V, VLJ, VLI) and our TAME. The backbone is ViT-B/16. The best results are boldfaced.}
\label{tab:clean}
\end{table}

\subsection{Experimental Setup}
\subsubsection{Datasets and Models}
We experiment on 11 benchmark datasets (ImageNet validation set and 10 other zero-shot test datasets): ImageNet~\cite{russakovsky2015imagenet}, Caltech101~\cite{fei2004learning}, DTD~\cite{cimpoi2014describing}, EuroSAT~\cite{helber2019eurosat}, Pets~\cite{parkhi2012cats}, Aircraft~\cite{maji2013fine}, Food101~\cite{bossard2014food}, Flowers~\cite{nilsback2008automated}, Cars~\cite{krause20133d}, SUN397~\cite{xiao2010sun}, and UCF101~\cite{soomro2012ucf101}. Our experiments focus on the CLIP model, specifically utilizing the ViT-B/16, ViT-B/32, and ViT-L/14 backbones. Following standard CLIP usage, we used hand-crafted prompts as textual inputs. For example, the prompt \texttt{"a photo of a <class>, a type of pet"} was applied for the Pets dataset.

\subsubsection{Attack Configuration}  
We evaluate the zero-shot adversarial robustness of CLIP under both white-box and black-box attack settings. Specifically, we use PGD-100~\cite{madry2017towards} and CW~\cite{carlini2017cw} as white-box attacks, DI~\cite{xie2019improving} as a black-box attack, and AutoAttack~\cite{croce2020reliable} as a stronger and more comprehensive evaluation. The hyperparameters of PGD, CW, and DI follow the default settings in the TorchAttacks library~\cite{kim2020torchattacks}. Consistent with~\cite{mao2023understanding}, we report results under perturbation budgets of $\epsilon = 1/255$, $2/255$, and $4/255$.

\subsubsection{Defense Configuration}
We compare TAME with representative train-time and test-time defenses. For train-time adversarial prompt tuning baselines, we follow the original settings~\cite{zhang2023adversarial,li2024one}, using PGD-2 with step size $\alpha=1/255$ and updating only prompts while freezing CLIP. We evaluate several prompt designs, including APT-T~\cite{zhang2023adversarial,li2024one}, APT-V, APT-VLJ, APT-VLI, FAP~\cite{zhou2024few}, and NAP-Tuning~\cite{zhang2026nap}. For test-time adversarial prompt tuning, we include R-TPT~\cite{sheng2025r} and TAPT~\cite{wang2025tapt}. We also compare with MMoP, which relies on a fixed prompt bank without updating the CLIP backbone.

\subsubsection{Implementation Details}  
For TAME, we initialize the defensive prompt with MMoP pretrained on ImageNet for 100 epochs, using a batch size of 32 and a learning rate of 0.035. By default, TAME is configured with 5 expert prompts with token-mean routing in the V-L Independent setting. For test-time robustness, we generate 255 augmented views per test sample via random resized crops and horizontal flips, resulting in 256 images including the original. From these 256 images, we select the top 10\% most confident predictions (with the lowest entropy) and compute the average entropy of their predicted probabilities. For adversarial-clean alignment, we pre-compute embedding statistics from the public dataset (ImageNet) using adversarial MMoP and standard MMoP, respectively. We then optimize the defensive prompts by minimizing a combined loss of multi-view entropy and adversarial-clean alignment using the AdamW optimizer, with a learning rate of $5 \times 10^{-4}$ and an alignment weigh of $\alpha = 0.5$. All experiments were conducted on an HPC cluster with a single NVIDIA H200 GPU.

\subsection{Main Results}
\label{sec:main_results}

Tables~\ref{tab:vitb16_final} and~\ref{tab:vitb32_final} report zero-shot adversarial robustness on 11 benchmark datasets under PGD, CW, DI, and AutoAttack using CLIP ViT-B/16 and ViT-B/32. We compare TAME with vanilla CLIP using hand-crafted prompts, train-time adversarial prompt tuning methods, and recent test-time defenses. Vanilla CLIP is highly vulnerable to adversarial perturbations, with its average accuracy dropping to 0.1\% and 0.0\% under AutoAttack on ViT-B/16 and ViT-B/32, respectively. Train-time APT methods improve robustness but remain limited in the zero-shot setting. For example, the strongest APT variant, APT-VLI, achieves only 17.2\% and 11.0\% average AutoAttack robustness on ViT-B/16 and ViT-B/32. More advanced train-time defenses such as FAP and NAP-Tuning also show limited transferability to unseen datasets. These results suggest that prompts optimized on a fixed training distribution are insufficient for open-world adversarial robustness, where both task semantics and attack patterns can vary substantially at test time.

TAME consistently achieves the highest average robustness across both backbones and all attacks. On ViT-B/16, it obtains 51.6\%, 55.0\%, 50.9\%, and 54.1\% under PGD, CW, DI, and AutoAttack. On ViT-B/32, it reaches 47.9\%, 50.0\%, 48.6\%, and 49.1\% under the same attacks. Compared with vanilla CLIP, TAME improves average AutoAttack robustness by 54.0\% on ViT-B/16 and 49.1\% on ViT-B/32. Compared with the strongest train-time baseline APT-VLI, the gains are 36.9\% and 38.1\%. TAME also outperforms R-TPT under all attacks on ViT-B/32 and further improves upon the strong TAPT baseline in average robustness. Beyond average performance, TAME shows clear advantages on large-scale and semantically diverse datasets. On ImageNet, it consistently improves over TAPT across both backbones and all attacks, with notable gains under PGD and AutoAttack. Similar trends are observed on fine-grained and diverse datasets such as Pets, Cars, and UCF101. These results indicate that the single-prompt formulation of TAPT, while effective, still has limited capacity to handle heterogeneous visual semantics and adversarial distortions. By dynamically composing an input-conditioned mixture of expert prompts, TAME provides more expressive and sample-specific defensive behavior, establishing a new state of the art for zero-shot adversarial robustness without requiring downstream training data or task-specific annotations.

\begin{figure*}[t!]
    \centering
    \includegraphics[width=1\linewidth]{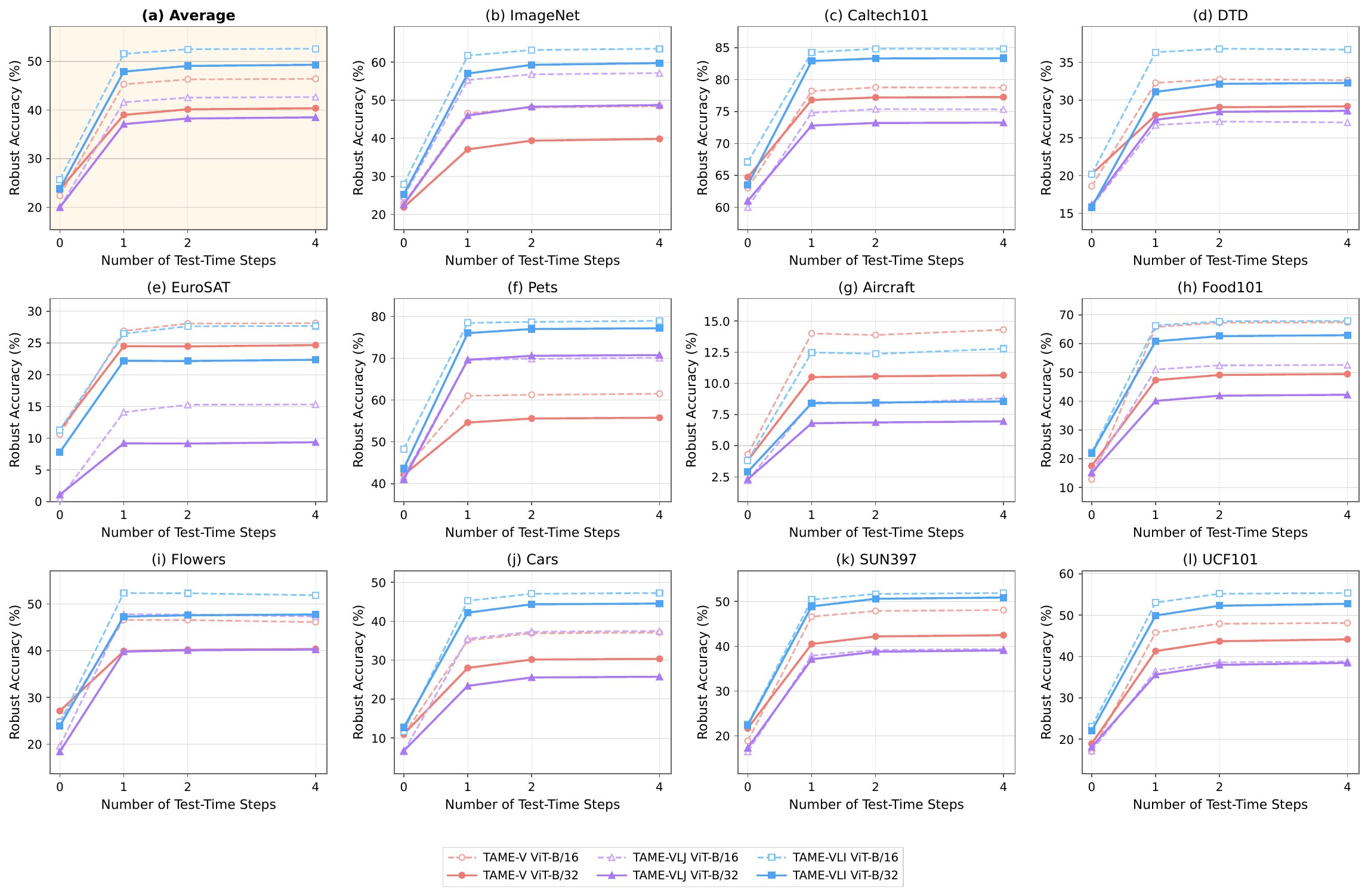}
   \caption{Adversarial robustness (\%) of our TAME method under different test-time robustness steps (i.e., \{0, 1, 2, 4\}). The results are reported against the PGD-100 attack on ViT-B/16 and ViT-B/32 architectures.}
    \label{fig4}
\end{figure*}

\subsection{Different Prompt Designs}
\label{sec:prompt_design}

\subsubsection{Zero-Shot Adversarial Robustness}
Table~\ref{tab:pgd_MMoP_tame_all} evaluates the generality of TAME across three representative prompt designs, including Visual Only, V-L Joint, and V-L Independent. For each design, we compare TAME with its MMoP counterpart, which adopts the same mixture-of-prompts parameterization but without test-time robustness. TAME consistently improves over MMoP across all prompt designs and CLIP backbones, demonstrating that our test-time robustness objective is not tied to a particular prompt parameterization. On ViT-B/16, TAME improves the average robust accuracy from 22.4\% to 45.3\% for V, from 20.0\% to 41.6\% for VLJ, and from 25.7\% to 51.6\% for VLI, yielding absolute gains of +22.9\%, +21.6\%, and +25.9\%, respectively. Similar improvements are observed on ViT-B/32 and ViT-L/14, indicating that the effectiveness of TAME remains stable across different model scales. Among the three prompt designs, VLI achieves the best robustness after applying TAME, reaching 51.6\%, 47.9\%, and 59.8\% average robust accuracy on ViT-B/16, ViT-B/32, and ViT-L/14, respectively. This indicates that independently modeling visual and textual prompt experts provides greater flexibility for adversarial correction than visual-only or V-L joint prompt. The improvements are also broad across datasets. For example, with VLI on ViT-B/16, TAME improves over MMoP by +33.7\% on ImageNet, +30.1\% on Pets, +44.1\% on Food101, and +33.6\% on Cars. These results demonstrate that TAME offers a general and effective mechanism for improving zero-shot adversarial robustness across prompt designs, model scales, and datasets.

\subsubsection{Zero-Shot Clean Accuracy}
Table~\ref{tab:clean} presents the zero-shot clean accuracy of different defense methods from ImageNet to downstream datasets. Overall, TAME preserves and often slightly improves clean performance compared with MMoP. The average clean accuracy increases from 64.4\% to 64.6\% for V, from 53.9\% to 54.3\% for VLJ, and from 63.0\% to 63.4\% for VLI. Under the strongest VLI design, TAME improves clean accuracy on most datasets, including ImageNet (+0.6\%), Caltech101 (+0.4\%), DTD (+0.6\%), EuroSAT (+0.6\%), Pets (+0.8\%), Cars (+0.6\%), SUN397 (+0.5\%), and UCF101 (+0.5\%), with only negligible drops on Aircraft and Flowers (-0.2\%). This indicates that TAME does not trade CLIP's zero-shot generalization for robustness. Instead, multi-view entropy minimization and adversarial-clean alignment encourage confident, robust, and semantically consistent adaptation. These results underscore TAME's ability to improve adversarial robustness without compromising too much clean accuracy.


\subsection{Number of Test-Time Steps}
\label{sec:steps_ablation}
We study the impact of the number of test-time steps in TAME. Figure~\ref{fig4} reports zero-shot adversarial robustness under PGD-100 with 0, 1, 2, and 4 optimization steps, where $\text{step}=0$ corresponds to directly using the initialized prompts without test-time adaptation. Introducing even a single adaptation step substantially improves robustness across datasets, backbones, and prompt designs. Specifically, on ViT-B/16 and ViT-B/32, one-step TAME improves the average zero-shot adversarial robustness by 21.6--25.9 and 15.3--24.1 percentage points, respectively, indicating that most of the robustness gain can be achieved with only one update. 
However, datasets vary in sensitivity to the number of TAME steps; for example, performance stabilizes quickly on datasets like EuroSAT, whereas for datasets such as ImageNet and UCF101, further improvements are observed with additional steps. The figure also illustrates that TAME’s benefits are consistent across different prompt designs.  We further measure the per-sample runtime overhead and find that TAME introduces only 0.105s, 0.172s, and 0.171s additional inference time for Visual-only, V-L Joint, and V-L Independent prompts, respectively, suggesting that one-step adaptation provides a favorable trade-off between robustness and efficiency.

\subsection{Different Perturbation Budgets}  
\begin{figure}[ht!]
    \centering
    \includegraphics[width=1\linewidth]{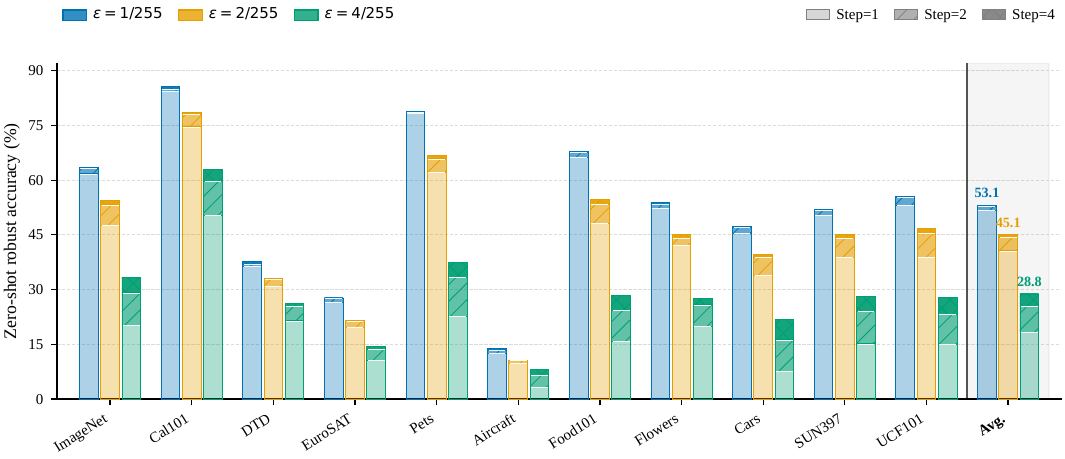}
    \caption{Zero-shot adversarial robustness (y-axis) ofs TAME under varying perturbation budgets $\epsilon$ (1/255, 2/255, and 4/255) and TAME steps (1, 2, and 4).}
    \label{fig5}
\end{figure}

We further evaluate TAME under different attack strengths by varying the perturbation budget $\epsilon$. Figure~\ref{fig5} reports the zero-shot adversarial accuracy across 11 datasets with $\epsilon \in \{1/255, 2/255, 4/255\}$ and different numbers of TAME adaptation steps. As expected, robust accuracy decreases as $\epsilon$ increases, indicating stronger adversarial attacks. Nevertheless, TAME consistently benefits from additional adaptation steps across all perturbation budgets. On average, increasing the adaptation steps from 1 to 4 improves robust accuracy from 51.6\% to 53.1\% for $\epsilon=1/255$, from 40.6\% to 45.1\% for $\epsilon=2/255$, and from 18.3\% to 28.8\% for $\epsilon=4/255$. The larger improvement under $\epsilon=4/255$ suggests that iterative test-time robustness is especially effective when the input suffers from stronger adversarial perturbations.

\subsection{Prompt Depth, Prompt Length, and Number of Experts}
\label{sec:depth_length_expert}
\begin{figure}[ht]
    \centering
    \includegraphics[width=1\linewidth]{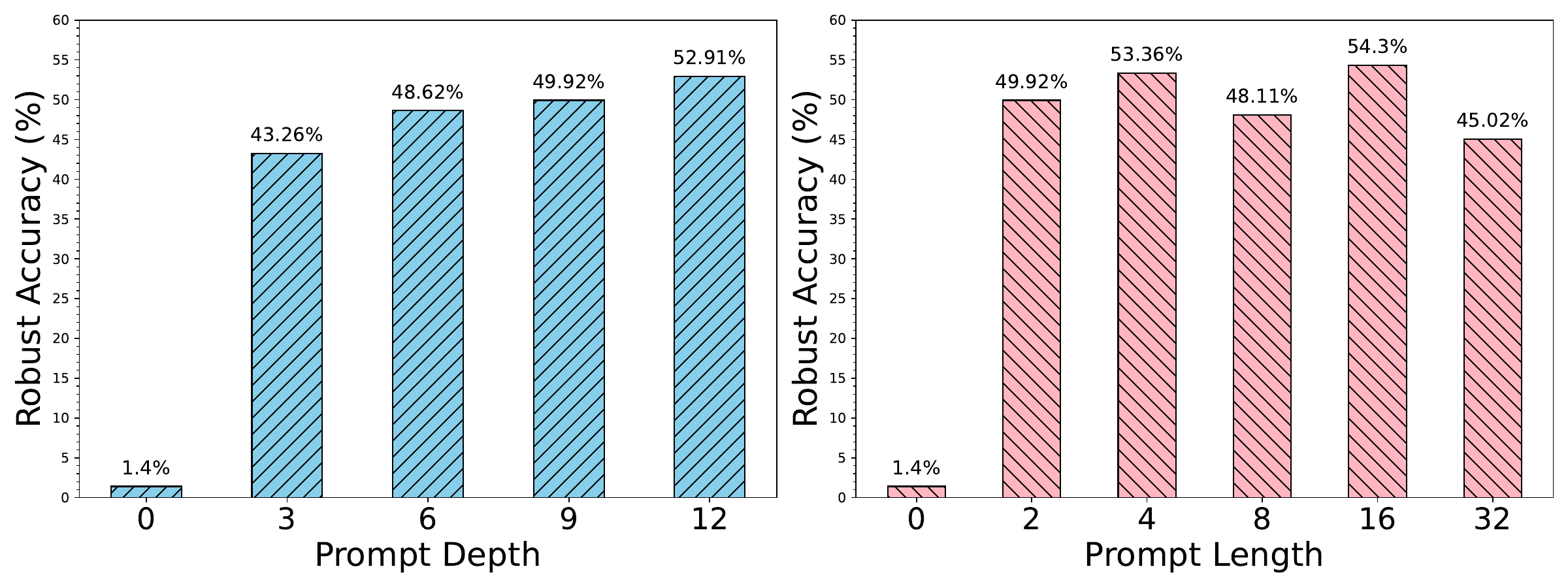}
    \caption{Effect of prompt depth and prompt length on ImageNet.}
    \label{fig6}
\end{figure}

We study the effects of prompt depth, prompt length, and the number of experts on zero-shot adversarial robustness. As shown in Figure~\ref{fig6}, increasing the prompt depth greatly improves robust accuracy on ImageNet, from 1.40\% at depth 0 to 43.26\% at depth 3 and 49.92\% at depth 9. Further increasing the depth to 12 yields a smaller gain, reaching 52.91\%, indicating diminishing returns with deeper prompts. Prompt length shows a similar trend: increasing the length from 0 to 2 improves robust accuracy from 1.40\% to 49.92\%, and length 4 further improves it to 53.36\%. However, the performance drops at larger lengths, e.g., 48.11\% at length 8 and 45.02\% at length 32, suggesting that overly long prompts may introduce redundant or unstable tokens. These results show that TAME benefits from sufficient prompt capacity, but overly large prompts do not necessarily improve robustness.

\begin{figure}[htbp]
    \centering
    \includegraphics[width=0.75\linewidth]{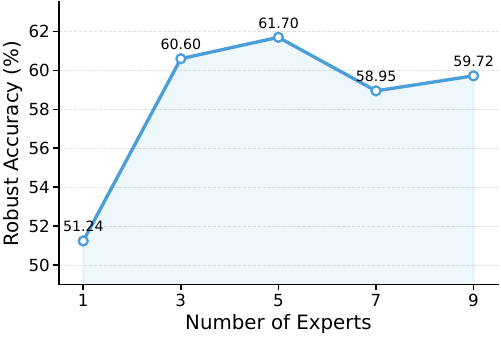}
    \caption{Effect of the number of experts on ImageNet.}
    \label{fig7}
\end{figure}

We further evaluate the effect of the number of prompt experts. As shown in Figure~\ref{fig7}, multiple experts substantially outperform a single expert, increasing robust accuracy from 51.24\% with one expert to 60.60\% with three experts and 61.70\% with five experts. Using more experts brings no further improvement, with accuracy dropping to 58.95\% and 59.72\% for seven and nine experts, respectively. This suggests that a moderate number of experts is sufficient to capture useful prompt variations for test-time robustness, while too many experts may introduce redundant choices or make optimization harder. Therefore, TAME achieves the best trade-off with compact prompt capacity and a moderate expert set.

\subsection{Effect of Alignment Layer Range}
\label{sec:align_layer}

\begin{figure}[htbp]
    \centering
    \includegraphics[width=0.8\linewidth]{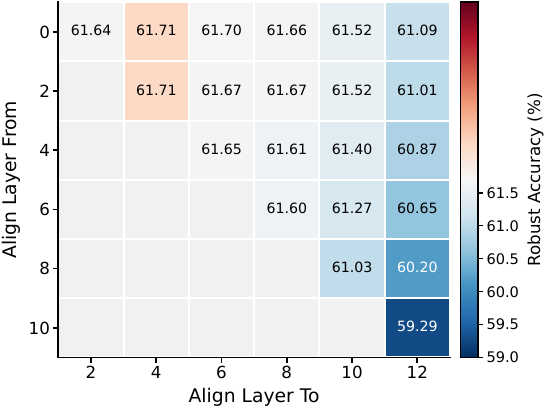}
    \caption{Effect of the alignment layer range. Each cell reports zero-shot robust accuracy for a specific alignment range.}
    \label{fig8}
\end{figure}

We evaluate the effect of the layer range used for feature alignment. As shown in Figure~\ref{fig8}, alignment over early-to-middle layers yields consistently strong robustness. The ranges 0--6, 0--4, and 2--4 achieve 61.70\%, 61.71\%, and 61.71\% robust accuracy on ImageNet, respectively. In contrast, alignment restricted to late layers performs worse, e.g., 60.20\% for 8--12 and 59.29\% for 10--12. Extending the range to all layers also provides no additional benefit, with 0--12 achieving 61.09\%. These results indicate that early-to-middle feature alignment is sufficient and more effective for robust test-time robustness.

\subsection{Effectiveness of TAME Loss} \label{ablation}
We conducte an ablation analysis on various components of TAME, specifically examining the impact of multi-view entropy minimization and adversarial-clean alignment. To isolate their effects, the MoE regularization is kept in all variants to stabilize expert routing. As shown in Table~\ref{tab:ablation}, the MMoP baseline achieves 25.31\% robust accuracy. Using only the entropy loss (TAME$^{\dag}$) improves the accuracy to 30.46\%, while using only the alignment loss (TAME$^{\ddag}$) further increases it to 57.63\%. Combining both losses yields the best performance of 61.70\%, showing that multi-view confidence optimization and adversarial-clean consistency provide complementary benefits for robust test-time adaptation.

\begin{table}[htbp]
    \centering
    \resizebox{0.95\linewidth}{!}{
    \begin{tabular}{lcccc}
    \toprule
    \textbf{Method}       & \textbf{Entropy Loss} & \textbf{Alignment Loss} & \textbf{Robust Accuracy} \\
    \midrule
    MMoP                  &               &               & 25.31\\
    $\text{TAME}^{\dag}$  & $\checkmark$  &               & 30.46\\
    $\text{TAME}^{\ddag}$ &               & $\checkmark$  & 57.63\\
    \midrule
    TAME                  & $\checkmark$  & $\checkmark$  & \textbf{61.70} \\
    \bottomrule
    \end{tabular}}
    \caption{Analysis of the impact of entropy and alignment losses. The average robust accuracy (\%) on ImageNet is reported. $\text{TAME}^{\dag}$ denotes our method excluding the adversarial-clean alignment. $\text{TAME}^{\ddag}$ denotes our method excluding the multi-view entropy loss.}
    \label{tab:ablation}
\end{table}

\subsection{Test-Time Reset Intervals}
\label{sec:reset_interval}

We investigate the effect of different reset intervals during test-time robustness. By default, TAME resets the learnable prompts before processing each test sample, preventing information leakage across samples and reducing the risk of adversarial poisoning. As shown in Table~\ref{tab:3}, increasing the reset interval from 1 to 32 brings a small but consistent improvement. For TAME, robust accuracy increases from 61.71\% with per-sample reset to 62.76\% with reset interval 32. A similar trend is observed for TAPT, improving from 49.92\% to 51.62\%. This suggests that adapting prompts over several consecutive samples can accumulate useful test-time information. However, removing the reset entirely leads to severe performance collapse, with accuracy dropping to 0.14\% for TAME and 0.48\% for TAPT. This indicates that unrestricted continual adaptation is highly unstable under adversarial inputs and may suffer from error accumulation or poisoning. Therefore, although larger reset intervals can provide marginal gains, we adopt the conservative per-sample reset strategy by default for a more reliable and attack-resilient test-time defense.

\begin{table}[ht]
    \centering
    \footnotesize
    \setlength{\tabcolsep}{4.0mm}
    \begin{tabular}{lcc}
    \toprule
    \textbf{Reset Interval} & TAPT & TAME \\
    \midrule
    reset=1   & 49.92          & 61.70          \\
    reset=2   & 50.20          & 61.93          \\
    reset=4   & 50.69          & 62.24          \\
    reset=8   & 51.14          & 62.51          \\
    reset=16  & 51.49          & 62.68          \\
    reset=32  & \textbf{51.62} & \textbf{62.76} \\
    reset=all & 0.48           & 0.14           \\
    \bottomrule
    \end{tabular}
    \caption{Zero-shot adversarial robustness (\%) of TAPT and TAME with varying reset intervals on ImageNet. ``reset=$N$'' means the prompt is reset after every $N$ test samples.}
    \label{tab:3}
\end{table}

\section{Limitation}
\label{sec:limit}
As a test-time defense method, TAME has certain limitations that warrant further research. Our method primarily addresses attacks in the image modality by dynamically adapting multimodal mixture-of-prompts and aligning image embeddings with pre-computed public data statistics. Future work could explore additional modality alignment and acceleration techniques to facilitate TAME's deployment in industrial applications. Moreover, our current focus is limited to image recognition tasks. Extending TAME to a broader range of tasks, such as visual reasoning and visual question answering in advanced models like GPT~\cite{openai2024gpt4} and Gemini~\cite{team2023gemini}, represents a promising direction for future research.

\section{Conclusion and Discussion}
\label{sec:conclusion}
In this paper, we propose a Test-Time Adversarial Mixture-of-Experts \textbf{TAME} framework for improving the zero-shot adversarial robustness of pre-trained VLMs. Unlike existing adversarial prompt tuning methods that rely on task-specific training data, TAME adapts mixture-of-prompts on the fly for each unlabeled test sample while keeping the VLM backbone frozen. By dynamically aggregating a bank of expert prompts through an input-conditioned router, TAME improves the expressive capacity and specialization ability of test-time robustness. In addition, TAME provides a unified formulation compatible with different prompt designs, including visual only, V-L joint, and V-L independent prompts. The adaptation is guided by multi-view entropy minimization, adversarial-clean distribution alignment, and MoE regularization, requiring no downstream annotations. Extensive experiments on 11 benchmarks and multiple CLIP backbones show that TAME consistently improves zero-shot adversarial robustness under various attacks while preserving clean accuracy. These results demonstrate that TAME is an effective and practical defense for robust open-world deployment of vision-language models. Future research could explore the scalability of test-time defense across different modalities.

\bibliographystyle{IEEEtran}
\bibliography{main}

@article{ma2026safety,
  title={Safety at scale: A comprehensive survey of large model and agent safety},
  author={Ma, Xingjun and Gao, Yifeng and Wang, Yixu and Wang, Ruofan and Wang, Xin and Sun, Ye and Ding, Yifan and Xu, Hengyuan and Chen, Yunhao and Zhao, Yunhan and others},
  journal={Foundations and Trends in Privacy and Security},
  year={2026}
}

@inproceedings{wang2025tapt,
  title={TAPT: Test-time adversarial prompt tuning for robust inference in vision-language models},
  author={Wang, Xin and Chen, Kai and Zhang, Jiaming and Chen, Jingjing and Ma, Xingjun},
  booktitle={CVPR},
  year={2025}
}

@article{wang2026openrt,
  title={OpenRT: An Open-Source Red Teaming Framework for Multimodal LLMs},
  author={Wang, Xin and Chen, Yunhao and Li, Juncheng and Wang, Yixu and Yao, Yang and Gu, Tianle and Li, Jie and Teng, Yan and Wang, Yingchun and Hu, Xia},
  journal={arXiv preprint arXiv:2601.01592},
  year={2026}
}

@article{min2022lossless,
  title={Lossless medical image compression based on anatomical information and deep neural networks},
  author={Min, Qiusha and Wang, Xin and Huang, Bo and Zhou, Zhongwei},
  journal={Biomedical Signal Processing and Control},
  year={2022}
}

@article{gao2025imperceptible,
  title={Imperceptible jailbreaking against large language models},
  author={Gao, Kuofeng and Li, Yiming and Du, Chao and Wang, Xin and Ma, Xingjun and Xia, Shu-Tao and Pang, Tianyu},
  journal={arXiv preprint arXiv:2510.05025},
  year={2025}
}

@article{chen2025evolve,
  title={Evolve the method, not the prompts: Evolutionary synthesis of jailbreak attacks on llms},
  author={Chen, Yunhao and Wang, Xin and Li, Juncheng and Wang, Yixu and Li, Jie and Teng, Yan and Wang, Yingchun and Ma, Xingjun},
  journal={arXiv preprint arXiv:2511.12710},
  year={2025}
}

@article{min2020web,
  title={Web-based technology for remote viewing of radiological images: App validation},
  author={Min, Qiusha and Wang, Xin and Huang, Bo and Xu, Liangzhou},
  journal={Journal of Medical Internet Research},
  year={2020}
}

@article{wang2025freezevla,
  title={Freezevla: Action-freezing attacks against vision-language-action models},
  author={Wang, Xin and Li, Jie and Weng, Zejia and Wang, Yixu and Gao, Yifeng and Pang, Tianyu and Du, Chao and Teng, Yan and Wang, Yingchun and Wu, Zuxuan and others},
  journal={arXiv preprint arXiv:2509.19870},
  year={2025}
}

@inproceedings{radford2021learning,
  title={Learning transferable visual models from natural language supervision},
  author={Radford, Alec and Kim, Jong Wook and Hallacy, Chris and Ramesh, Aditya and Goh, Gabriel and Agarwal, Sandhini and Sastry, Girish and Askell, Amanda and Mishkin, Pamela and Clark, Jack and others},
  booktitle={ICML},
  year={2021}
}

@inproceedings{jia2021scaling,
  title={Scaling up visual and vision-language representation learning with noisy text supervision},
  author={Jia, Chao and Yang, Yinfei and Xia, Ye and Chen, Yi-Ting and Parekh, Zarana and Pham, Hieu and Le, Quoc and Sung, Yun-Hsuan and Li, Zhen and Duerig, Tom},
  booktitle={ICML},
  year={2021}
}

@inproceedings{zhang2023multi,
  title={Multi-event video-text retrieval},
  author={Zhang, Gengyuan and Ren, Jisen and Gu, Jindong and Tresp, Volker},
  booktitle={ICCV},
  year={2023}
}

@article{huang2023visual,
  title={A visual--language foundation model for pathology image analysis using medical twitter},
  author={Huang, Zhi and Bianchi, Federico and Yuksekgonul, Mert and Montine, Thomas J and Zou, James},
  journal={Nature Medicine},
  year={2023}
}

@inproceedings{wang2022medclip,
  title={MedCLIP: Contrastive Learning from Unpaired Medical Images and Text},
  author={Wang, Zifeng and Wu, Zhenbang and Agarwal, Dinesh and Sun, Jimeng},
  booktitle={EMNLP},
  year={2022}
}

@article{ahn2022can,
  title={Do as i can, not as i say: Grounding language in robotic affordances},
  author={Ahn, Michael and Brohan, Anthony and Brown, Noah and Chebotar, Yevgen and Cortes, Omar and David, Byron and Finn, Chelsea and Fu, Chuyuan and Gopalakrishnan, Keerthana and Hausman, Karol and others},
  journal={preprint arXiv:2204.01691},
  year={2022}
}

@inproceedings{shridhar2022cliport,
  title={Cliport: What and where pathways for robotic manipulation},
  author={Shridhar, Mohit and Manuelli, Lucas and Fox, Dieter},
  booktitle={CoRL},
  year={2022}
}

@inproceedings{khandelwal2022simple,
  title={Simple but effective: Clip embeddings for embodied ai},
  author={Khandelwal, Apoorv and Weihs, Luca and Mottaghi, Roozbeh and Kembhavi, Aniruddha},
  booktitle={CVPR},
  year={2022}
}

@inproceedings{szegedy2013intriguing,
  title={Intriguing properties of neural networks},
  author={Szegedy, Christian and Zaremba, Wojciech and Sutskever, Ilya and Bruna, Joan and Erhan, Dumitru and Goodfellow, Ian and Fergus, Rob},
  booktitle={ICLR},
  year={2013}
}

@inproceedings{madry2017towards,
  title={Towards deep learning models resistant to adversarial attacks},
  author={Madry, Aleksander and Makelov, Aleksandar and Schmidt, Ludwig and Tsipras, Dimitris and Vladu, Adrian},
  booktitle={ICLR},
  year={2018}
}

@inproceedings{dong2018boosting,
  title={Boosting adversarial attacks with momentum},
  author={Dong, Yinpeng and Liao, Fangzhou and Pang, Tianyu and Su, Hang and Zhu, Jun and Hu, Xiaolin and Li, Jianguo},
  booktitle={CVPR},
  year={2018}
}

@inproceedings{zhao2024evaluating,
  title={On evaluating adversarial robustness of large vision-language models},
  author={Zhao, Yunqing and Pang, Tianyu and Du, Chao and Yang, Xiao and Li, Chongxuan and Cheung, Ngai-Man Man and Lin, Min},
  booktitle={NeurIPS},
  year={2024}
}

@inproceedings{zhang2022towards,
  title={Towards adversarial attack on vision-language pre-training models},
  author={Zhang, Jiaming and Yi, Qi and Sang, Jitao},
  booktitle={ACM MM},
  year={2022}
}

@inproceedings{zhou2023advclip,
  title={Advclip: Downstream-agnostic adversarial examples in multimodal contrastive learning},
  author={Zhou, Ziqi and Hu, Shengshan and Li, Minghui and Zhang, Hangtao and Zhang, Yechao and Jin, Hai},
  booktitle={ACM MM},
  year={2023}
}

@article{ma2024imbalanced,
  title={Imbalanced gradients: a subtle cause of overestimated adversarial robustness},
  author={Ma, Xingjun and Jiang, Linxi and Huang, Hanxun and Weng, Zejia and Bailey, James and Jiang, Yu-Gang},
  journal={Machine Learning},
  year={2024}
}

@inproceedings{croce2020reliable,
  title={Reliable evaluation of adversarial robustness with an ensemble of diverse parameter-free attacks},
  author={Croce, Francesco and Hein, Matthias},
  booktitle={ICML},
  year={2020}
}

@inproceedings{zhang2023adversarial,
  title={Adversarial prompt tuning for vision-language models},
  author={Zhang, Jiaming and Ma, Xingjun and Wang, Xin and Qiu, Lingyu and Wang, Jiaqi and Jiang, Yu-Gang and Sang, Jitao},
  booktitle={ECCV},
  year={2024}
}

@inproceedings{li2024one,
  title={One prompt word is enough to boost adversarial robustness for pre-trained vision-language models},
  author={Li, Lin and Guan, Haoyan and Qiu, Jianing and Spratling, Michael},
  booktitle={CVPR},
  year={2024}
}

@inproceedings{zhou2024few,
  title={Few-Shot Adversarial Prompt Learning on Vision-Language Models},
  author={Zhou, Yiwei and Xia, Xiaobo and Lin, Zhiwei and Han, Bo and Liu, Tongliang},
  booktitle={NeurIPS},
  year={2024}
}

@inproceedings{zhang2019theoretically,
  title={Theoretically principled trade-off between robustness and accuracy},
  author={Zhang, Hongyang and Yu, Yaodong and Jiao, Jiantao and Xing, Eric and El Ghaoui, Laurent and Jordan, Michael},
  booktitle={ICML},
  year={2019}
}

@inproceedings{su2018robustness,
  title={Is robustness the cost of accuracy?--a comprehensive study on the robustness of 18 deep image classification models},
  author={Su, Dong and Zhang, Huan and Chen, Hongge and Yi, Jinfeng and Chen, Pin-Yu and Gao, Yupeng},
  booktitle={ECCV},
  year={2018}
}

@article{pedraza2021relationship,
  title={On the relationship between generalization and robustness to adversarial examples},
  author={Pedraza, Anibal and Deniz, Oscar and Bueno, Gloria},
  journal={Symmetry},
  year={2021}
}

@article{bahng2022exploring,
  title={Exploring visual prompts for adapting large-scale models},
  author={Bahng, Hyojin and Jahanian, Ali and Sankaranarayanan, Swami and Isola, Phillip},
  journal={preprint arXiv:2203.17274},
  year={2022}
}

@inproceedings{shu2022test,
  title={Test-time prompt tuning for zero-shot generalization in vision-language models},
  author={Shu, Manli and Nie, Weili and Huang, De-An and Yu, Zhiding and Goldstein, Tom and Anandkumar, Anima and Xiao, Chaowei},
  booktitle={NeurIPS},
  year={2022}
}

@inproceedings{abdul2024align,
  title={Align your prompts: Test-time prompting with distribution alignment for zero-shot generalization},
  author={Abdul Samadh, Jameel and Gani, Mohammad Hanan and Hussein, Noor and Khattak, Muhammad Uzair and Naseer, Muhammad Muzammal and Shahbaz Khan, Fahad and Khan, Salman H},
  booktitle={NeurIPS},
  year={2024}
}

@inproceedings{wang2024revisiting,
  title={Revisiting Adversarial Training at Scale},
  author={Wang, Zeyu and Li, Xianhang and Zhu, Hongru and Xie, Cihang},
  booktitle={CVPR},
  year={2024}
}

@inproceedings{mao2023understanding,
  title={Understanding Zero-shot Adversarial Robustness for Large-Scale Models},
  author={Chengzhi Mao and Scott Geng and Junfeng Yang and Xin Wang and Carl Vondrick},
  booktitle={ICLR},
  year={2023}
}

@InProceedings{schlarmannrobust,
  title = {Robust {CLIP}: Unsupervised Adversarial Fine-Tuning of Vision Embeddings for Robust Large Vision-Language Models},
  author = {Schlarmann, Christian and Singh, Naman Deep and Croce, Francesco and Hein, Matthias},
  booktitle = {ICML},
  year = {2024}
}

@inproceedings{wang2024pre,
  title={Pre-trained model guided fine-tuning for zero-shot adversarial robustness},
  author={Wang, Sibo and Zhang, Jie and Yuan, Zheng and Shan, Shiguang},
  booktitle={CVPR},
  year={2024}
}

@article{zhou2024revisiting,
  title={Revisiting the Adversarial Robustness of Vision Language Models: a Multimodal Perspective},
  author={Zhou, Wanqi and Bai, Shuanghao and Zhao, Qibin and Chen, Badong},
  journal={preprint arXiv:2404.19287},
  year={2024}
}

@inproceedings{wang2024advqdet,
  title={Adv{QD}et: Detecting Query-Based Adversarial Attacks with Adversarial Contrastive Prompt Tuning},
  author={Xin Wang and Kai Chen and Xingjun Ma and Zhineng Chen and Jingjing Chen and Yu-Gang Jiang},
  booktitle={ACM MM},
  year={2024}
}

@article{russakovsky2015imagenet,
  title={Imagenet large scale visual recognition challenge},
  author={Russakovsky, Olga and Deng, Jia and Su, Hao and Krause, Jonathan and Satheesh, Sanjeev and Ma, Sean and Huang, Zhiheng and Karpathy, Andrej and Khosla, Aditya and Bernstein, Michael and others},
  journal={IJCV},
  year={2015}
}

@inproceedings{fei2004learning,
  title={Learning generative visual models from few training examples: An incremental bayesian approach tested on 101 object categories},
  author={Fei-Fei, Li and Fergus, Rob and Perona, Pietro},
  booktitle={CVPR Workshops},
  year={2004}
}

@inproceedings{parkhi2012cats,
  title={Cats and dogs},
  author={Parkhi, Omkar M and Vedaldi, Andrea and Zisserman, Andrew and Jawahar, CV},
  booktitle={CVPR},
  year={2012}
}

@inproceedings{cimpoi2014describing,
  title={Describing textures in the wild},
  author={Cimpoi, Mircea and Maji, Subhransu and Kokkinos, Iasonas and Mohamed, Sammy and Vedaldi, Andrea},
  booktitle={CVPR},
  year={2014}
}

@article{helber2019eurosat,
  title={Eurosat: A novel dataset and deep learning benchmark for land use and land cover classification},
  author={Helber, Patrick and Bischke, Benjamin and Dengel, Andreas and Borth, Damian},
  journal={IEEE J-STARS},
  year={2019}
}

@article{maji2013fine,
  title={Fine-grained visual classification of aircraft},
  author={Maji, Subhransu and Rahtu, Esa and Kannala, Juho and Blaschko, Matthew and Vedaldi, Andrea},
  journal={preprint arXiv:1306.5151},
  year={2013}
}

@inproceedings{bossard2014food,
  title={Food-101--mining discriminative components with random forests},
  author={Bossard, Lukas and Guillaumin, Matthieu and Van Gool, Luc},
  booktitle={ECCV},
  year={2014}
}

@inproceedings{nilsback2008automated,
  title={Automated flower classification over a large number of classes},
  author={Nilsback, Maria-Elena and Zisserman, Andrew},
  booktitle={ICVGIP},
  year={2008},
}

@inproceedings{krause20133d,
  title={3d object representations for fine-grained categorization},
  author={Krause, Jonathan and Stark, Michael and Deng, Jia and Fei-Fei, Li},
  booktitle={ICCV Workshops},
  year={2013}
}

@inproceedings{xiao2010sun,
  title={Sun database: Large-scale scene recognition from abbey to zoo},
  author={Xiao, Jianxiong and Hays, James and Ehinger, Krista A and Oliva, Aude and Torralba, Antonio},
  booktitle={CVPR},
  year={2010}
}

@article{soomro2012ucf101,
  title={UCF101: A dataset of 101 human actions classes from videos in the wild},
  author={Soomro, K},
  journal={preprint arXiv:1212.0402},
  year={2012}
}

@inproceedings{xie2019improving,
  title={Improving transferability of adversarial examples with input diversity},
  author={Xie, Cihang and Zhang, Zhishuai and Zhou, Yuyin and Bai, Song and Wang, Jianyu and Ren, Zhou and Yuille, Alan L},
  booktitle={CVPR},
  year={2019}
}

@article{kim2020torchattacks,
  title={Torchattacks: A pytorch repository for adversarial attacks},
  author={Kim, Hoki},
  journal={preprint arXiv:2010.01950},
  year={2020}
}

@article{schneider2020improving,
  title={Improving robustness against common corruptions by covariate shift adaptation},
  author={Schneider, Steffen and Rusak, Evgenia and Eck, Luisa and Bringmann, Oliver and Brendel, Wieland and Bethge, Matthias},
  journal={NeurIPS},
  year={2020}
}

@article{nado2020evaluating,
  title={Evaluating prediction-time batch normalization for robustness under covariate shift},
  author={Nado, Zachary and Padhy, Shreyas and Sculley, D and D'Amour, Alexander and Lakshminarayanan, Balaji and Snoek, Jasper},
  journal={preprint arXiv:2006.10963},
  year={2020}
}

@inproceedings{wangtent,
  title={Tent: Fully Test-Time Adaptation by Entropy Minimization},
  author={Wang, Dequan and Shelhamer, Evan and Liu, Shaoteng and Olshausen, Bruno and Darrell, Trevor},
  booktitle={ICLR},
  year={2021}
}

@inproceedings{zhang2022memo,
  title={Memo: Test time robustness via adaptation and augmentation},
  author={Zhang, Marvin and Levine, Sergey and Finn, Chelsea},
  booktitle={NeurIPS},
  year={2022}
}

@inproceedings{wang2022continual,
  title={Continual test-time domain adaptation},
  author={Wang, Qin and Fink, Olga and Van Gool, Luc and Dai, Dengxin},
  booktitle={CVPR},
  year={2022}
}

@inproceedings{niu2022efficient,
  title={Efficient test-time model adaptation without forgetting},
  author={Niu, Shuaicheng and Wu, Jiaxiang and Zhang, Yifan and Chen, Yaofo and Zheng, Shijian and Zhao, Peilin and Tan, Mingkui},
  booktitle={ICML},
  year={2022}
}

@inproceedings{gan2020large,
  title={Large-scale adversarial training for vision-and-language representation learning},
  author={Gan, Zhe and Chen, Yen-Chun and Li, Linjie and Zhu, Chen and Cheng, Yu and Liu, Jingjing},
  booktitle={NeurIPS},
  year={2020}
}

@article{team2023gemini,
  title={Gemini: a family of highly capable multimodal models},
  author={Team, Gemini and Anil, Rohan and Borgeaud, Sebastian and Alayrac, Jean-Baptiste and Yu, Jiahui and Soricut, Radu and Schalkwyk, Johan and Dai, Andrew M and Hauth, Anja and Millican, Katie and others},
  journal={arXiv:2312.11805},
  year={2023}
}

@inproceedings{lu2023set,
  title={Set-level guidance attack: Boosting adversarial transferability of vision-language pre-training models},
  author={Lu, Dong and Wang, Zhiqiang and Wang, Teng and Guan, Weili and Gao, Hongchang and Zheng, Feng},
  booktitle={ICCV},
  year={2023}
}

@article{he2023sa,
  title={Sa-attack: Improving adversarial transferability of vision-language pre-training models via self-augmentation},
  author={He, Bangyan and Jia, Xiaojun and Liang, Siyuan and Lou, Tianrui and Liu, Yang and Cao, Xiaochun},
  journal={preprint arXiv:2312.04913},
  year={2023}
}

@inproceedings{wang2024transferable,
  title={Transferable multimodal attack on vision-language pre-training models},
  author={Wang, Haodi and Dong, Kai and Zhu, Zhilei and Qin, Haotong and Liu, Aishan and Fang, Xiaolin and Wang, Jiakai and Liu, Xianglong},
  booktitle={IEEE S\&P},
  year={2024}
}

@inproceedings{yin2024vlattack,
  author = {Yin, Ziyi and Ye, Muchao and Zhang, Tianrong and Du, Tianyu and Zhu, Jinguo and Liu, Han and Chen, Jinghui and Wang, Ting and Ma, Fenglong},
  title = {VLATTACK: Multimodal Adversarial Attacks on Vision-Language Tasks via Pre-trained Models},
  booktitle = {NeurIPS},
  year = {2023}
}

@inproceedings{zhang2024universal,
  title={Universal Adversarial Perturbations for Vision-Language Pre-trained Models},
  author={Zhang, Peng-Fei and Huang, Zi and Bai, Guangdong},
  booktitle={ACM SIGIR},
  pages={862--871},
  year={2024}
}

@article{fang2024one,
  title={One Perturbation is Enough: On Generating Universal Adversarial Perturbations against Vision-Language Pre-training Models},
  author={Fang, Hao and Kong, Jiawei and Yu, Wenbo and Chen, Bin and Li, Jiawei and Xia, Shutao and Xu, Ke},
  journal={preprint arXiv:2406.05491},
  year={2024}
}

@inproceedings{zanella2024test,
  title={On the test-time zero-shot generalization of vision-language models: Do we really need prompt learning?},
  author={Zanella, Maxime and Ben Ayed, Ismail},
  booktitle={CVPR},
  year={2024}
}

@inproceedings{hussein2024promptsmooth,
  title={Promptsmooth: Certifying robustness of medical vision-language models via prompt learning},
  author={Hussein, Noor and Shamshad, Fahad and Naseer, Muzammal and Nandakumar, Karthik},
  booktitle={MICCAI},
  year={2024}
}

@inproceedings{luo2024adversarial,
  title={Adversarial prompt distillation for vision-language models},
  author={Luo, Lin and Wang, Xin and Zi, Bojia and Zhao, Shihao and Ma, Xingjun and Jiang, Yu-Gang},
  booktitle={ICASSP},
  year={2026}
}

@article{wang2023exploring,
  title={Exploring transferability of multimodal adversarial samples for vision-language pre-training models with contrastive learning},
  author={Wang, Youze and Hu, Wenbo and Dong, Yinpeng and Zhang, Hanwang and Su, Hang and Hong, Richang},
  journal={preprint arXiv:2308.12636},
  year={2023}
}

@inproceedings{fan2024mixprompt,
  title={MixPrompt: Enhancing Generalizability and Adversarial Robustness for Vision-Language Models via Prompt Fusion},
  author={Fan, Hao and Ma, Zhaoyang and Li, Yong and Tian, Rui and Chen, Yunli and Gao, Chenlong},
  booktitle={ICIC},
  year={2024}
}

@inproceedings{xing2025clip,
  title={Clip is strong enough to fight back: Test-time counterattacks towards zero-shot adversarial robustness of clip},
  author={Xing, Songlong and Zhao, Zhengyu and Sebe, Nicu},
  booktitle={CVPR},
  year={2025}
}

@inproceedings{sheng2025r,
  title={R-TPT: Improving Adversarial Robustness of Vision-Language Models through Test-Time Prompt Tuning},
  author={Sheng, Lijun and Liang, Jian and Wang, Zilei and He, Ran},
  booktitle={CVPR},
  year={2025}
}

@article{jacobs1991adaptive,
  title={Adaptive mixtures of local experts},
  author={Jacobs, Robert A and Jordan, Michael I and Nowlan, Steven J and Hinton, Geoffrey E},
  journal={Neural Computation},
  year={1991}
}

@article{du2025mixture,
  title={Mixture of prompts learning for vision-language models},
  author={Du, Yu and Niu, Tong and Zhao, Rong},
  journal={Frontiers in Artificial Intelligence},
  year={2025}
}

@inproceedings{choi2023smop,
  title={Smop: Towards efficient and effective prompt tuning with sparse mixture-of-prompts},
  author={Choi, Joon-Young and Kim, Junho and Park, Jun-Hyung and Mok, Wing-Lam and Lee, SangKeun},
  booktitle={EMNLP},
  year={2023}
}

@article{chen2022plot,
  title={Plot: Prompt learning with optimal transport for vision-language models},
  author={Chen, Guangyi and Yao, Weiran and Song, Xiangchen and Li, Xinyue and Rao, Yongming and Zhang, Kun},
  journal={preprint arXiv:2210.01253},
  year={2022}
}

@inproceedings{bulat2023lasp,
  title={Lasp: Text-to-text optimization for language-aware soft prompting of vision \& language models},
  author={Bulat, Adrian and Tzimiropoulos, Georgios},
  booktitle={CVPR},
  year={2023}
}

@article{wang2024one,
  title={One Prompt is not Enough: Automated Construction of a Mixture-of-Expert Prompts},
  author={Wang, Ruochen and An, Sohyun and Cheng, Minhao and Zhou, Tianyi and Hwang, Sung Ju and Hsieh, Cho-Jui},
  journal={preprint arXiv:2407.00256},
  year={2024}
}

@article{zhao2025enhancing,
  title={Enhancing Adversarial Robustness of Vision Language Models via Adversarial Mixture Prompt Tuning},
  author={Zhao, Shiji and Zhu, Qihui and Xiong, Shukun and Ruan, Shouwei and Fan, Yize and Duan, Ranjie and Guo, Qing and Wei, Xingxing},
  journal={preprint arXiv:2505.17509},
  year={2025}
}

@inproceedings{shazeer2017outrageously,
  title={Outrageously large neural networks: The sparsely-gated mixture-of-experts layer},
  author={Shazeer, Noam and Mirhoseini, Azalia and Maziarz, Krzysztof and Davis, Andy and Le, Quoc and Hinton, Geoffrey and Dean, Jeff},
  booktitle={ICLR},
  year={2017}
}

@article{fedus2022switch,
  title={Switch transformers: Scaling to trillion parameter models with simple and efficient sparsity},
  author={Fedus, William and Zoph, Barret and Shazeer, Noam},
  journal={JMLR},
  year={2022}
}

@inproceedings{carlini2017cw,
  title={Towards evaluating the robustness of neural networks},
  author={Carlini, Nicholas and Wagner, David},
  booktitle={IEEE S\&P},
  year={2017}
}

@article{zhang2026nap,
  title={NAP-Tuning: Neural Augmented Prompt Tuning for Adversarially Robust Vision-Language Models},
  author={Zhang, Jiaming and Wang, Xin and Ma, Xingjun and Qiu, Lingyu and Jiang, Yu-Gang and Sang, Jitao},
  journal={IEEE Transactions on Pattern Analysis and Machine Intelligence},
  year={2026}
}

@article{openai2024gpt4,
  title={GPT-4 technical report},
  author={Achiam, Josh and Adler, Steven and Agarwal, Sandhini and Ahmad, Lama and Akkaya, Ilge and Aleman, Florencia Leoni and Almeida, Diogo and Altenschmidt, Janko and Altman, Sam and Anadkat, Shyamal and others},
  journal={arXiv preprint arXiv:2303.08774},
  year={2023}
}

\end{document}